# Machine Learning Co-pilot for Screening of Organic Molecular Additives for Perovskite Solar Cells

Yang Pu[1,#], Zhiyuan Dai[1,#], Yifan Zhou[1], Ning Jia[1], Hongyue Wang[1], Yerzhan Mukhametkarimov[2], Ruihao Chen[1*], Hongqiang Wang[1,*], Zhe Liu[1,*]

[1]School of Materials Science and Engineering, State Key Laboratory of Solidification Processing, Northwestern Polytechnical University, 127 Youyi West Road, Xi'an, Shaanxi, P.R. China 710072

[2] Department of Physics and Technology, Al-Farabi Kazakh National University, 71 Al-Farabi Avenue, Almaty, Republic of Kazakhstan, 050040

[#]These authors contributed equally to this work.

[*]Correspondence authors. E-mails: zhe.liu@nwpu.edu.cn (Z. Liu); hongqiang.wang@nwpu.edu.cn (H.-Q. Wang); rhchen@nwpu.edu.cn (R. Chen).

## Abstract

Machine learning (ML) has been extensively employed in planar perovskite photovoltaics to screen effective organic molecular additives, while encountering predictive biases for novel materials due to small datasets and reliance on predefined descriptors. Present work thus proposes an effective approach, Co-Pilot for Perovskite Additive Screener (Co-PAS), an ML-driven framework designed to accelerate additive screening for perovskite solar cells (PSCs). Co-PAS overcomes predictive biases by integrating the Molecular Scaffold Classifier (MSC) for scaffold-based pre-screening and utilizing Junction Tree Variational Autoencoder (JTVAE) latent vectors to enhance molecular structure representation, thereby enhancing the accuracy of power conversion efficiency (PCE) predictions. Leveraging Co-PAS, we integrate domain knowledge to screen an extensive dataset of 250,000 molecules from PubChem, prioritizing candidates based on predicted PCE values and key molecular properties such as donor number, dipole moment, and hydrogen bond acceptor count. This workflow leads to the identification of several promising passivating molecules, including the novel Boc-L-threonine N-hydroxysuccinimide ester (BTN), which, to our knowledge, has not been explored as an additive in PSCs and achieves a device PCE of 25.20%. Our results underscore the potential of Co-PAS in advancing additive discovery for high-performance PSCs.

# 1.Introduction

Perovskite solar cells (PSCs) have rapidly emerged as one of the most promising candidates for next-generation cost-effective photovoltaics, owing to their exceptional optical and electronic properties, including wide absorption range, high carrier mobility, long carrier diffusion length, and low exciton binding energy[1]. Despite significant advancements, defects in perovskite film absorbers—comprising both point defects (such as vacancies, interstitials, and anti-site defects) and extended defects (such as surface and grain boundaries)[2]—inevitably lead to nonradiative recombination, hindering further improvements in efficiency[3]. Leveraging organic molecule additives as defect passivators has demonstrated considerable potential in enhancing PSCs performance[2]. The organic additive can be either incorporated during the perovskite deposition steps (i.e., adding in precursor or antisolvent)[4] or applied in the post-processing step onto the deposited perovskite thin films[5]. By employing these integration methods, organic additives in PSCs have been shown to reduce recombination centers[6], stabilize lattice structures[7], and passivate defects including surface traps[8] and iodine vacancies[9], resulting in improved device performance. Various passivation mechanisms, including ionic passivation through coordination bonds with defect vacancies[10], lone pair electron passivation by Lewis acid-base coordination from functional groups[11], and hydrogen bond passivation using hydrogen atoms near electronegative atoms to form intermolecular hydrogen bonds[12], have been reported to enhance defect passivation. Nonetheless, the effective selection of passivating materials is critical, as it relies on the specific functional groups and molecular structure, which determine their ability to interact with defects at surfaces and grain boundaries, thereby enhancing passivation performance[13]. However, this process is challenged by the vast chemical space of potential candidates.

Thoroughly exploring the chemical space poses substantial challenges, as traditional experimental screening requires extensive time and effort to accurately assess the potential of molecular candidates effectively. For example, over 110 million organic small molecules exist in the PubChem database[14], while other molecular databases, such as ChEMBL[15], ZINC[16], and MOSES[17], collectively house millions more. This sheer volume of organic molecule databases complicates systematic exploration. Machine learning (ML) has recently been attempted to accelerate material discovery and uncover underlying relationships between material characteristics and properties[18]. Given the limited experimental data on additives, the development of advanced molecular representation techniques is crucial for improving model accuracy and accelerating discoveries. Recent studies have made progress by quantitatively analyzing molecular characteristics that drive improvements in PSCs and light emitting diodes.[19]. For example, ML algorithms were systematically used to analyze the relationship between molecular descriptors and device power conversion efficiency (PCE), enabling the discovery of promising candidates as additives, such as Phenylpropan-1-aminium iodide (2-PPAI), 2-(p-tolyl) ethan-1-aminium iodide (4-Me-PEAI), and phenylpropan-1-aminium iodide (PPAI), for enhanced device efficiencies[20][21]. To improve predictive performance, researchers have integrated molecular fingerprints, including the Molecular Access System (MACCS), FP2, and Extended-connectivity Fingerprints (ECFP), with molecular descriptors to capture essential structural and chemical information[22]. However, these approaches face limitations. ML models trained on small, imbalanced datasets (which are based on less than two hundred molecules in the PSC publications) often struggle to generalize predictions to unseen molecules with diverse structures (which are up to a million available organic molecules), leading to prediction inaccuracies or even worse cases of random guesses. On the other hand, predefined descriptors and molecular fingerprints, which rely on domain-specific rules, may fail to capture critical structural nuances and introduce biases. As such, efficient methodologies for molecule representation, especially data-driven methodologies, are essential for enhancing model general applicability in additive screening for PSCs. Addressing these challenges requires improved data analytics strategies for model training and prediction.

In this study, we proposed Co-Pilot for Perovskite Additive Screener (Co-PAS), an ML-based tool designed to screen candidate additives for enhancing the performance of PSCs. A small dataset of 129 molecules, reported as effective passivating additives, was curated to provide a representative foundation for ML-based analysis. Recognizing that limited dataset size could introduce significant predication deviation with unseen molecular structures, we first developed the Molecular Scaffold Classifier (MSC). This classifier partitions the training and test sets based on molecular scaffolds, ensuring the model sees and learns from the full variety of scaffold types present in the molecule dataset. To improve prediction accuracy further, we incorporated the encoded latent vectors from Junction Tree Variational Autoencoder (JTVAE) as inputs for the ML model, which predict the PCE of PSCs with additives. With the optimized model, we identified dozens of promising additive candidates and integrated domain knowledge to refine the screening process with chemical property criteria relevant to PCE, including donor number, dipole moment, and hydrogen bond acceptor. This screening process led to the identification of two standout additives—Risedronic acid (RIS) and Boc-L-threonine N-hydroxysuccinimide ester (BTN)—that exhibited high PCE. The experimental validation confirmed the effectiveness of these additives, achieving PCEs of 24.38% for RIS and 25.20% for BTN. These findings highlight the potential of Co-PAS as a convenient and powerful tool for advancing the discovery and optimization of PSC additives, paving the way for higher perovskite device efficiency.

# 2. Results and Discussion

## 2.1. Machine Learning Workflow

**Figure 1a** illustrates the systematic workflow of the Co-pilot for Perovskite Additive Screening (Co-PAS), a tool designed to identify promising additive molecules for PSCs. This workflow consists of five key stages: (1) classification of molecular scaffolds, (2) prediction of PCE, (3) molecular property screening, (4) literature analysis for validation, and (5) experimental assessment of candidate additives.

To build the foundation for Co-PAS, we generated the scaffolds of the molecules in our dataset using the open-source software package RDKit[23]. These scaffolds were categorized into nine distinct groups based on structural similarities (detailed in **Figure S1**). In the first stage of the workflow, newly input molecules are evaluated to determine whether their scaffolds are already represented in the existing dataset. Molecules with known scaffolds proceed to the subsequent stages, while those with novel scaffolds are flagged for possible experimental exploration, which might contribute to the expansion of the dataset. This stage is to mimic human experts' way to conduct molecular additive screening and prevent the random guesses in the predictions when the model has not seen any similar molecule structures.

In the second stage, PCE prediction is performed using the established regression model to identify top-performing candidates. To ensure robust predictions, we partitioned the dataset by randomly selecting one or two molecules from each scaffold group for the test set, while the remaining molecules formed for training, maintaining a consistent 9:1 training-to-test ratio. Molecules were represented using the canonical Simplified Molecular Input Line Entry System (SMILES), a standardized format for molecule description[24]. As illustrated in **Figure 1b**, we employed two key molecular representations: MACCS fingerprints and RDKit molecular descriptors. MACCS fingerprints provide concise structural encoding through a fixed-length fingerprint[25]. Meanwhile, RDKit molecular descriptors quantitatively represent molecular properties[26]. Alongside these representations, the JTVAE-latent vectors can also serve as inputs. This representation allows us to uncover correlations between molecular additives and PCE performance.

Recent research studies have shown that molecular properties, such as donor number and dipole moment, play a significant role in improving photovoltaic PCE in PSCs[21, 27]. Based on these findings, the third stage implements a systematic property screening process for the selected molecules. Here, each molecule is evaluated against predefined thresholds established based on domain expertise and prior research. This step ensures that the molecules exhibit the essential properties required to meet the rigorous performance criteria for PSCs, thereby refining the selection of high-performing candidates.

In the fourth stage, an extensive literature analysis is conducted. By incorporating scaffold-based classification, this workflow cross-references the selected candidates with previously studied compounds. Molecules with similar scaffolds and functional groups are identified, establishing connections between the screened additives and known substances. This step provides a robust foundation for subsequent experimental validation.

Finally, in the fifth stage, the candidate additives undergo experimental validation to assess their photovoltaic PCEs. Several key factors are meticulously considered to optimize experimental validation. For example, the solubility of each additive in the perovskite precursor solution is assessed to ensure uniform distribution and effective interaction. If an additive is not soluble, we would then consider applying it via post-treatment. These factors, informed by prior research, guide the experimental design and ensure a thorough assessment of the additives' effectiveness.

In summary, Co-PAS integrates molecular scaffold classification, PCE prediction, property screening, literature comparison, and experimental validation into a unified framework. Therefore, Co-PAS is used to the following work to demonstrate its potential for rapid screening and evaluation of additive molecules.

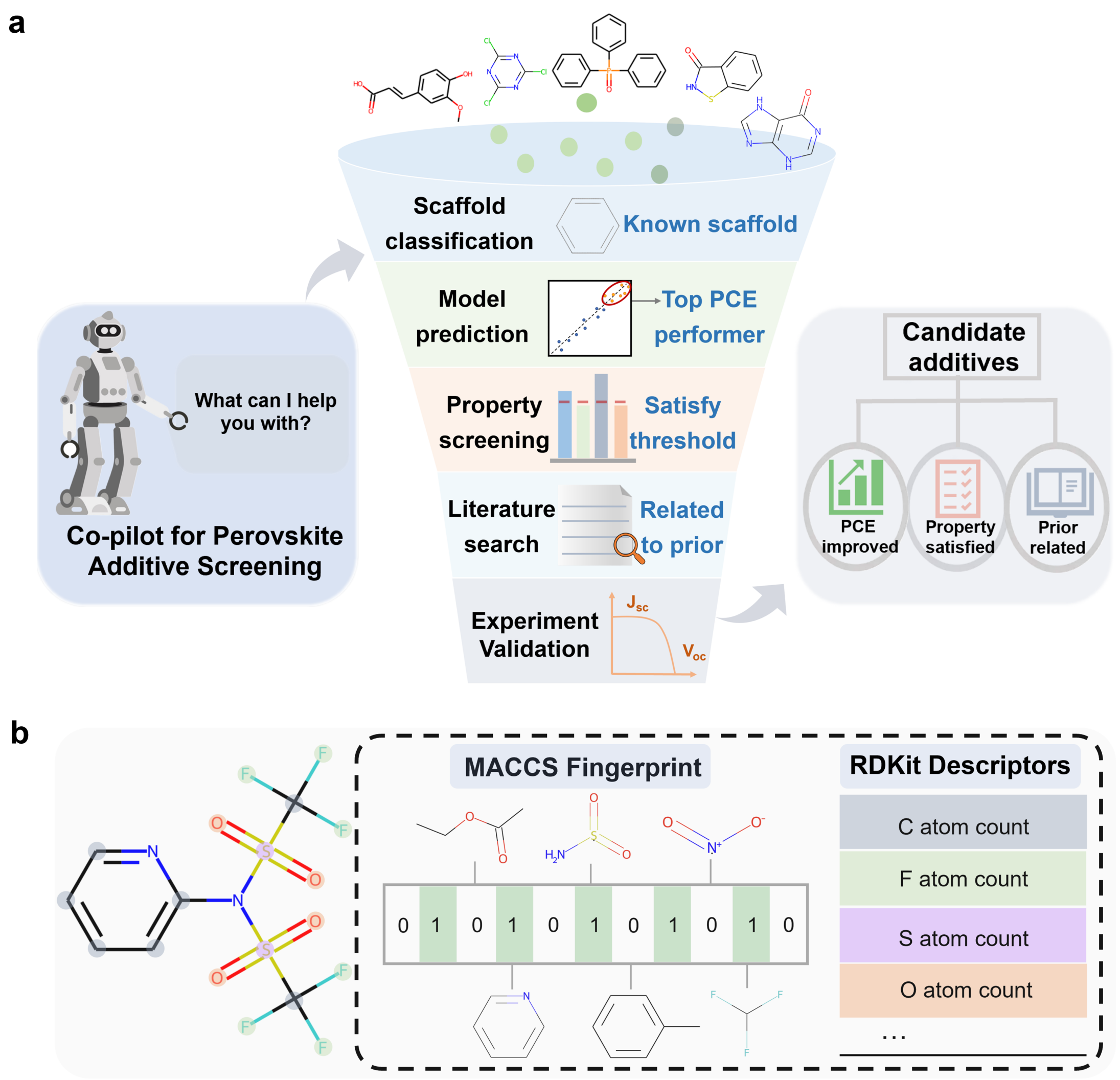

**Figure 1.** **(a)** Schematic of the Co-PAS workflow for screening candidate additives in PSCs. The screening process involves five steps: molecule scaffold classification, model prediction, property screening, literature search and experiment validation. **(b)** Methods for molecular representation, including MACCS fingerprints and RDKit molecular descriptors: MACCS fingerprints illustrate the binary encoding of molecular structures, RDKit molecular descriptors are used to quantify various molecular features.

## 2.2. Molecular Scaffold Classifier for Dataset Partitioning

Small datasets, such as the limited collection of perovskite passivation molecules, often lead to high variance and inherent bias in model performance when evaluated with traditional random splitting method[28]. These limitations ultimately hinder the model's ability to generalize and predict the performance of unseen molecules accurately. To overcome these obstacles, we proposed MSC, a novel data partitioning method that classifies molecules based on their scaffolds, thereby enhancing model accuracy. As illustrated in **Figure 2a** and **Figure S1**, the scaffolds were categorized into nine distinct groups according to their structural similarities, enabling us to effectively capture the underlying structural information of the molecules. This classification serves as a foundation for the analysis of molecular features and their impact on PCE of PSCs.

Building on this scaffold-based classification, we employed MACCS fingerprints and RDKit descriptors to explore the relationships between molecular features and the PCE of PSCs. With 208 descriptors generated through the RDKit tool, the high dimensionality of the input features posed challenges for accurate modeling, necessitating the removal of redundant features. To address this issue, we calculated Pearson correlation coefficients (PCC) to evaluate the linear relationships between descriptor pairs[29]. In line with established practices, we applied a PCC threshold of 0.9, retaining one representative descriptor for any pair exceeding this threshold[20]. This process reduced the original set of 208 descriptors to 44, effectively eliminating redundancy while preserving relevant features (detailed in Supporting Information, **Figure S2-S4**). The selected molecular descriptors, together with MACCS fingerprints and their combinations, were subsequently used as inputs for the regression model to predict the PCE of PSCs treated with additives. We opted for the Gradient Boosting model due to its capability to sequentially correct errors and effectively handle non-linear relationships between features[30]. To ensure a robust evaluation of the model's performance, we repeated this process 200 times and averaged the resulting metrics, providing a more comprehensive and reliable assessment of accuracy.

To assess the effectiveness of the MSC method, we conducted a thorough comparison with the traditional random splitting approach. The resulting Mean Absolute Errors (MAEs) and their corresponding deviations for both the MSC method and random splitting are shown in **Figure 2b** and **Table S1**. The MSC method notably outperformed random splitting, achieving a lower MAE with reduced deviation, demonstrating its improved precision in model evaluation. To further validate the effectiveness of the MSC method, we applied a Leave-One-Group-Out (LOGO) strategy by training on eight groups and testing on the remaining one group. This approach demonstrated that the MSC method consistently achieved narrower assessment errors compared to LOGO, confirming its robustness and accuracy. As a critical measure of the rank-based relationship between predicted and actual values, the Spearman correlation coefficient (SCC) provides insight into the alignment of the model's predictions with true data trends[31]. Thus, in addition to prediction error analysis, we also employed SCC to gain deeper insights into the model's performance. As illustrated in **Figure 2c** and **Table S1**, the MSC method achieved a notably higher SCC value with reduced deviation compared to both random split and LOGO. These results further underscore the superior effectiveness of the MSC method in enhancing the accuracy of machine learning model evaluations.

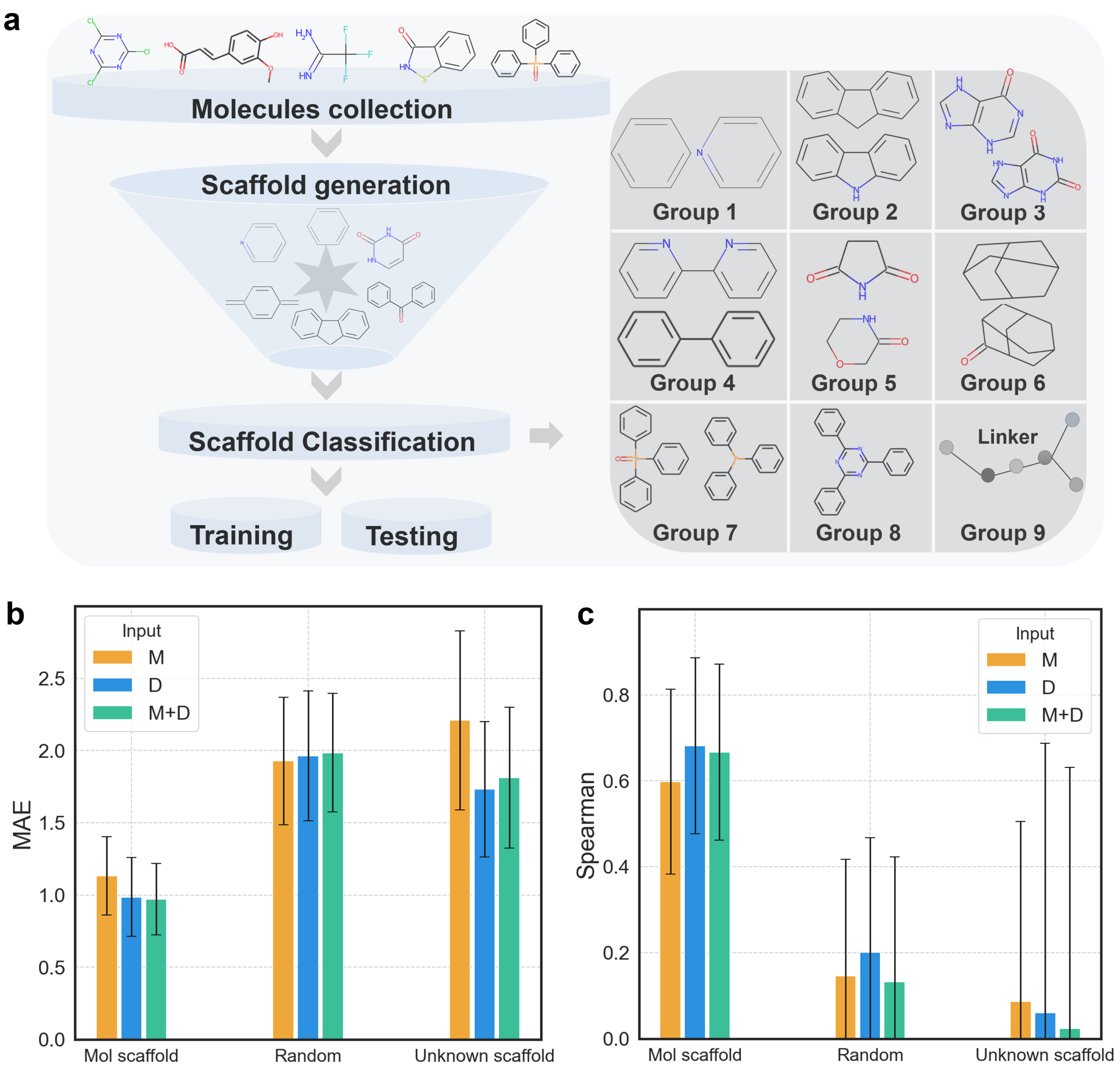

**Figure 2.** **(a)** Overview of the Molecule Scaffold Classifier (MSC) method. The MSC method classify additives based on their molecular scaffolds before partitioning them into training and testing datasets. Comparison of the **(b)** prediction accuracy and **(c)** Spearman coefficient for MSC method, random split, and unknown scaffold (M: MACCS, D: RDKit Descriptors).

## 2.3. Machine-learning Model Training and Analysis Incorporating JTVAE-Latent Vectors

Choice of an appropriate molecular representation is crucial for developing accurate models. The traditional approach of the MACCS fingerprint provides a simplified binary encoding based on specific structural fragments[25], but often fails to capture critical structural details, limiting its effectiveness. Molecular descriptors offer a quantitative view of chemical and physical properties through manually selected features[26]; however, their reliance on data quality can introduce biases and inaccuracies, especially when the data is insufficient or poorly curated. These shortcomings have spurred the exploration of more advanced molecular representation methods. Recent advances in structural learning and encoding, such as Variational Autoencoders (VAEs), including Character VAE (CVAE)[32], Grammar VAE (GVAE)[33], and Syntax-directed VAE (SDVAE)[34], utilize deep learning to map complex data distributions into a continuous latent space, enabling more nuanced molecular representations. Building on these developments, Junction Tree Variational Autoencoder (JTVAE) integrates tree-structured scaffold and graphical information to encode molecular structures more effectively. [35]. This approach provides a richer and more precise representation of molecular features and structural relationships of different atoms, capturing the rationality and validity of molecular structures. By leveraging JTVAE-latent vectors as model inputs, researchers can achieve significantly improved predictive performance due to a comprehensive understanding of molecular features. Despite the notable success of VAEs in drug development, their application to PSCs remains underexplored. Molecular characteristics are critical to PSC performance, such as charge transport and defect passivation. The arrangement of atoms and the presence of functional groups play a vital role in determining interactions within perovskite and molecular additive. By utilizing JTVAE-latent vectors, these intricate structural features can be effectively captured, enabling the design of innovative additives.

JTVAE represents a data-driven method in molecular representation compared to traditional methods, such as MACCS fingerprints and molecular descriptors. Trained on large molecular databases containing millions of compounds, JTVAE leverages continuous and smooth latent space, allowing for a more precise and comprehensive representation of molecular information through latent space vectors **(Figure S5-S7)**. The model employs a junction tree framework that integrates both the tree-structured scaffold and detailed molecular features, effectively capturing both structural and graphical information **(Figure 3a)**. This framework enables JTVAE to first capture the tree-structured scaffold (the junction tree), followed by a detailed representation of atoms and bonds, ensuring a highly accurate molecular expression **(Figure S8-S9)**. In addition to improving the accuracy of molecular representations, JTVAE enhances the exploration of chemical space, thereby facilitating the design of more diverse and innovative molecules **(Figure S10)**.

To investigate the effectiveness of JTVAE-latent vectors in improving model accuracy, we utilized several popular regression algorithms, including Random Forest (RF) regression, Gradient Boosting (GB) regression, and Support Vector Machine (SVM) regression[18c, 18d, 36]. We selected various molecular representations, such as MACCS molecular fingerprints, RDKit molecular descriptors, JTVAE-latent vectors, and their combinations as inputs for the model. The output of the model was the device PCE with the additive incorporation. We then established regression models with optimized hyperparameters to analyze the correlation between the additives and the PCEs. The hyperparameters of these models are shown in **Table S2**.

We employed the MSC approach for data partitioning and performed a thorough assessment of ML regression models. The full list of MAEs for these models with various model inputs can be found in **Table S3**. **Figure 3b** demonstrates that the use of latent vectors as inputs reduces the Mean Absolute Error (MAE). This finding suggests that JTVAE-latent vectors are effective in capturing comprehensive molecular features and improving the regression model prediction. Additionally, combining different representations, such as molecular descriptors with latent vectors, yielded lower error than using either representation alone. Notably, the GB model achieved the lowest prediction error, with the MAE of $0.8297 \pm 0.2477$. Therefore, we selected this model for subsequent screening of additive molecules.

We further validated the reliability of our model using 24 additional additive molecules from the literature **(Table S4)** as a validation set. As shown in **Figure 3c**, the comparison between the predicted outcomes from our model and the experimental results reveals a good agreement, confirming the predictive accuracy of our ML model for the unseen molecules. This validation result demonstrates the effectiveness of our model's predictions and highlights its potential to expedite the screening of high-performance passivation materials. One should note that all these 24 unseen molecules can pass the scaffold simililarity test by our MSC method described in section 2.2.

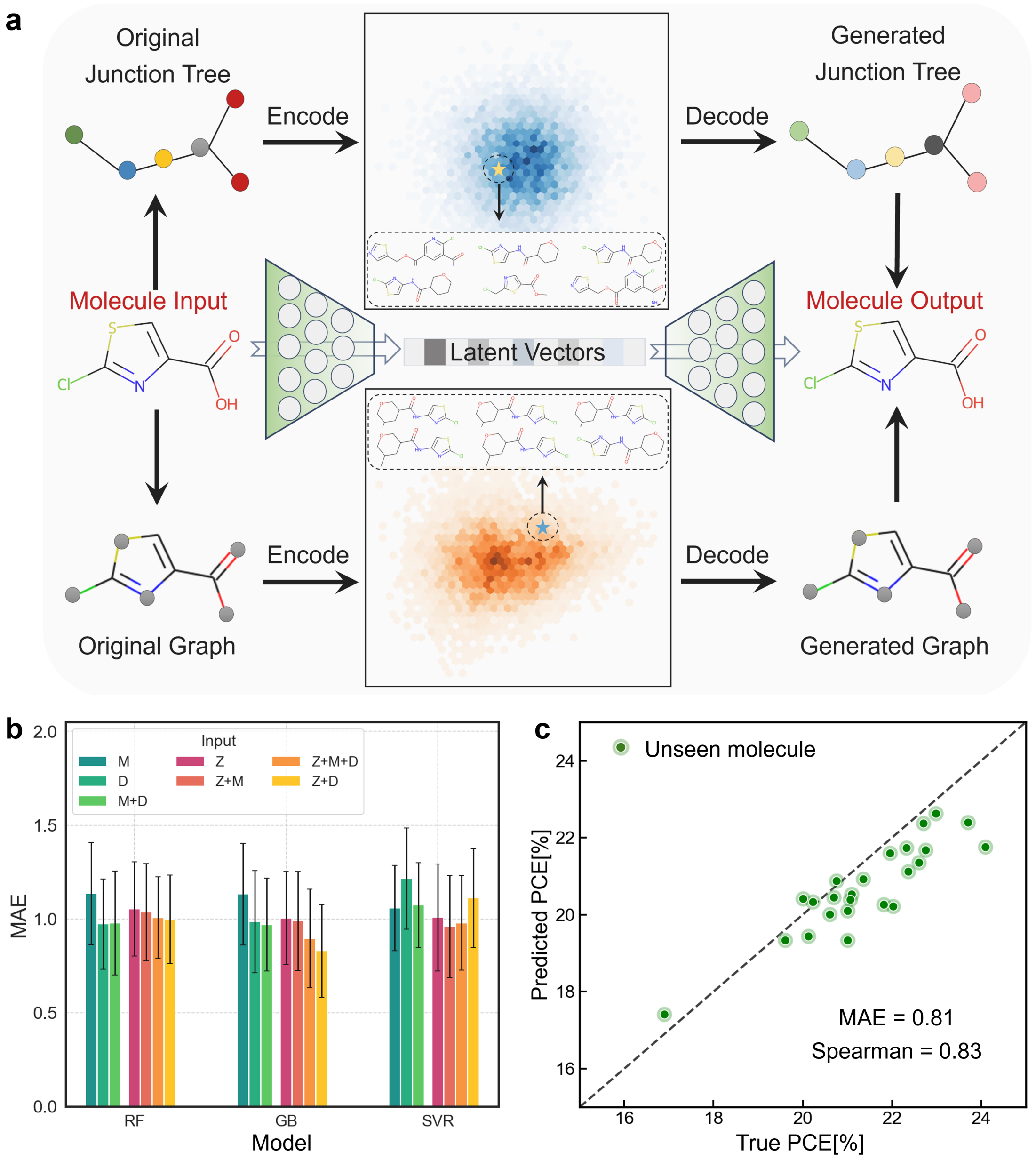

**Figure 3. (a)** The JTVAE encodes molecules as a junction tree of substructures, where each colored node represents a substructure, along with its corresponding graph. Both the tree and graph are encoded into their respective latent vectors. To reconstruct the molecule, the junction tree is first regenerated from its latent vectors, followed by the assembly of the tree's nodes to recreate the original molecular structure. **(b)** MAE values of machine learning regression models with multiple inputs based on MSC method. (M: MACCS, D: RDKit descriptors, Z: latent vector from JTVAE). **(c)** Comparison of the predicted PCE for the validation dataset from in the recently published papers, which is not included in the train and test datasets.

## 2.4. Screening New Molecules and Experimental Validation

Building on our established model for accurately predicting device PCE, we can rapidly evaluate the potential of individual molecules. Given the vast number of available organic small molecules, experimentally assessing the PCE of each candidate is infeasible. To address this challenge, we initially screened 250,000 randomly selected molecules from the PubChem database **(Figure 4a)**. Our model employs JTVAE-latent vectors as inputs, so molecules containing elements absent from JTVAE's cluster label vocabulary were excluded. To further refine our screening, we employed the MSC method to group molecules by scaffold. Molecules with scaffolds not represented in our dataset were removed, yielding a more refined dataset. Using this dataset of more than forty-seven thousand molecules, we predicted the PCE of the remaining candidates and focused on the top 1% with the highest predicted PCE values. Despite this rigorous screening process, 478 molecules remained, necessitating additional refinement for experimental feasibility.

To prioritize molecules with strong passivation potential, we integrated domain knowledge to evaluate candidate molecules based on three critical properties for defect passivation: Gutmann's donor number (DN), dipole moment (DM), and hydrogen bond acceptor count (HA). Defects at perovskite interfaces and grain boundaries can be passivated by Lewis bases, which donate lone pairs of electrons to undercoordinated Pb ions[13b, 37]. This interaction, which reduces carrier recombination and enhances perovskite stability, is quantified by Gutmann's donor number (DN), reflecting the molecule's Lewis basicity[27a]. Higher DN values indicate stronger coordination with $Pb^{2+}$ centers, inhibiting iodide binding and slowing perovskite crystallization. Additionally, dipole moment (DM) quantifies the separation of positive and negative charges within a molecule, serving as an indicator of its overall polarity[38]. Higher DM values enhance a molecule's electron-donating ability and increase its binding energy with uncoordinated $Pb^{2+}$ through coordinate bonds, thereby strengthening the passivation effect[27b, 39]. Furthermore, the hydrogen bond acceptor count (HA) is critical for enhancing the interaction between passivating molecules and defect sites within the perovskite lattice[40]. HA indicates the number of sites within a molecule capable of accepting hydrogen bonds. A higher HA increases binding site availability for hydrogen bond donors ($FA^+$ or $MA^+$), promoting stronger hydrogen bonding and enhancing the stability of the passivation layer[2]. By applying thresholds for these properties from previous studies[21, 27a], we identified 41 promising candidates, 30 of which had CAS codes.

To provide a solid foundation for experimental validation, we conducted a comprehensive scaffold-based analysis using the existing dataset, comparing the selected candidates with previously studied compounds. Representative molecules and their references are shown in **Figure 4b**, with a complete list provided in **Table S5**. This review highlights the importance of analogous scaffolds and functional groups in influencing PCE performance, linking the screened additives to established compounds.

Considering accessibility and cost, we selected three representative molecules—Risedronic acid (RIS), Boc-L-threonine N-hydroxysuccinimide ester (BTN), and NSC-2805 (NSC)—from the screened molecule list for experimental validation.

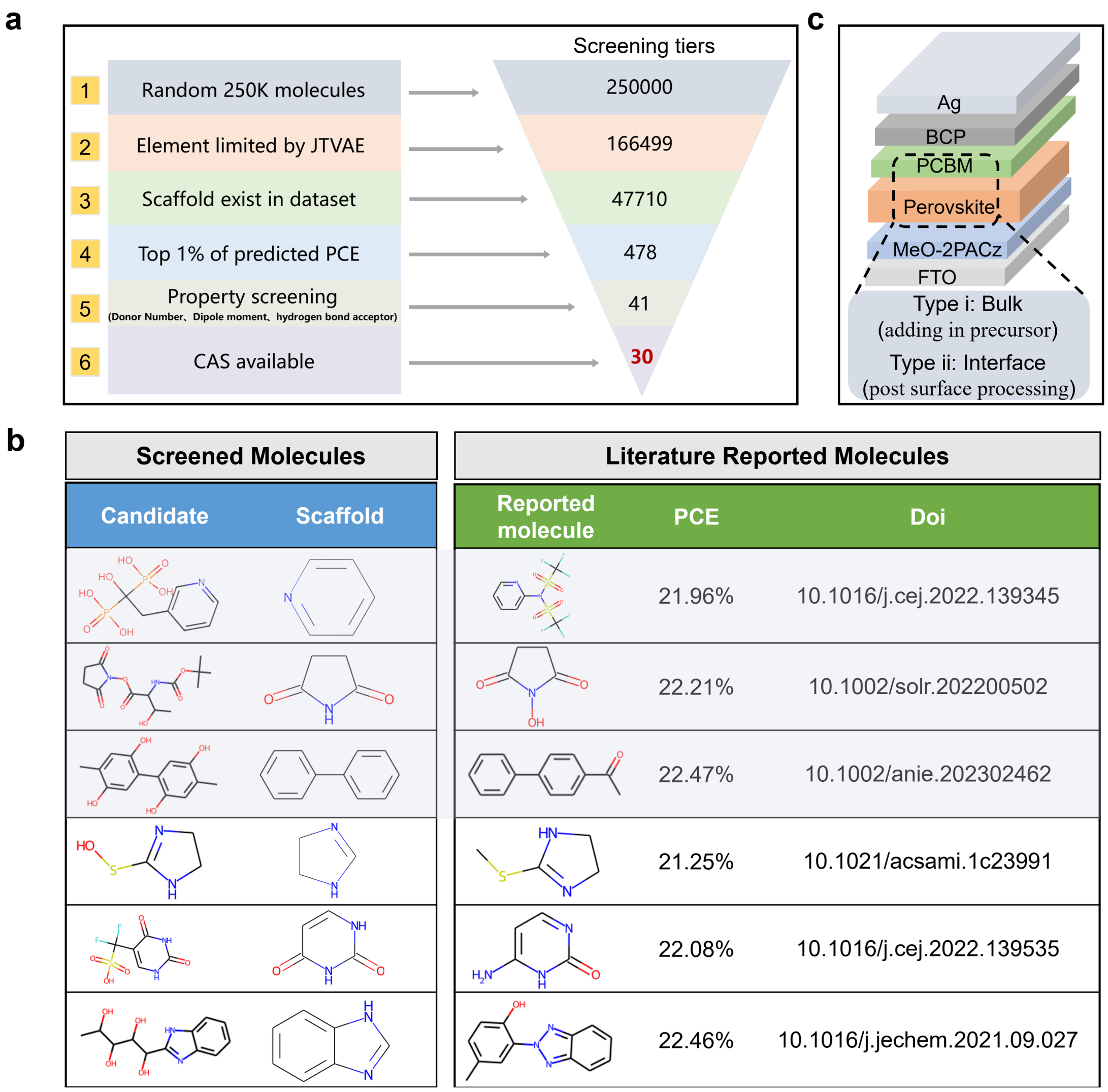

**Figure 4. (a)** Workflow of the material screening process, consisting of five screening tiers. **(b)** Representative candidate additives identified by Co-PAS, alongside relevant compounds from prior literature, with molecules shaded in gray selected for experimental validation. **(c)** Device configuration of the FAMACs-based PSCs used for experimental validation, along with the method of additive incorporation, including introducing into the precursor and modifying the interface.

To ensure rigorous experimental validation, we systematically evaluated the effects of additive incorporation both in the precursor solution and at the perovskite film interface **(Figure 4c)**. The application strategy was determined by the solubility of the selected additives in dimethylformamide (DMF) and isopropanol (IPA). RIS, insoluble in DMF but sparingly soluble in IPA, was applied exclusively at the perovskite/ electron transport layer (ETL) interface. BTN and NSC, soluble in both solvents, were applied both at the interface and within the precursor solution **(Figure 5a** and **Figure S11)**. For subsequent analyses, we focused on bulk incorporation for BTN and NSC due to their better performance in device PCEs. Given the excellent stability and advantages of low-temperature processing of PSCs with an inverted (*p-i-n*) structure[41], we fabricated a series of *p-i-n* PSCs with a composition of $FA_{0.85}MA_{0.10}Cs_{0.05}PbI_3$ to evaluate the photovoltaic performance after additive modification. The device architecture, shown in **Figure 4c**, consisted of FTO/MeO-2PACz/perovskite/PCBM/BCP/Ag. In *p-i-n* structure, the perovskite active layer generates photogenerated electron-hole pairs by absorbing sunlight, and the interface between the perovskite active layer and ETL serves as a crucial pathway for carrier transport[42]. Thus, optimizing the perovskite active layer and its interface with the ETL was prioritized in this study to enhance device performance.

**Figure 5b** summarizes the device performance for three different additives, while **Figure 5c** presents the *J-V* curves under reverse scanning conditions. We fabricated 24 PSCs and the narrow PCE distribution indicates good reproducibility. Devices treated with RIS and BTN achieved champion PCEs of 24.38% and 25.20%, respectively, demonstrating their potential in enhancing photovoltaic performance. In contrast, NSC-treated devices exhibited a reduced PCE of 22.72%, compared to the control devices with a PCE of 23.11%, which was attributed to inferior interfacial properties. Photovoltaic parameter analysis **(Table S6)** revealed that NSC-treated devices exhibited lower fill factor (FF) and open-circuit voltage ($V_{OC}$) compared to the control. The reduced FF indicates poor interface contact, while the lower $V_{OC}$ suggests an energy level mismatch at the perovskite/ETL interface. These combined deficiencies significantly compromised the overall efficiency of NSC-treated devices. To verify the reliability of short-circuit current ($J_{SC}$), external quantum efficiency (EQE) measurements were conducted (**Figure 5d**). The short-circuit current densities ($J_{SC}$) obtained were 24.56, 24.94, 25.03, and 24.43 mA $cm^{-2}$ for the control, RIS-, BTN-, and NSC-treated devices, respectively, consistent with the $J_{SC}$ values from the *J-V* curve (**Figure S12**).

To further elucidate the performance differences, time-resolved photoluminescence (TRPL) measurements were performed on perovskite films **(Figure 5e)**. The average carrier lifetimes ($\tau_{ave}$) increased from 249.98 ns for the control film to 343.44 ns and 627.64 ns for the RIS- and BTN-treated films, respectively, indicating reduced non-radiative recombination. In contrast, the NSC-treated films exhibited no improvement in carrier lifetime, consistent with their poorer performance. To evaluate the impact of additives on device performance, the space-charge-limited current (SCLC) method was employed to analyze trap-state densities in electron-only devices with a configuration of FTO/$SnO_2$/perovskite/PCBM/Ag **(Figure 5f)**. From the dark *J-V* curves, the trap-filled limit voltages (VTFL) were determined to be 0.59 V, 0.43 V, 0.37 V, and 0.61 V for the control, RIS-treated, BTN-treated, and NSC-treated devices, respectively. The corresponding trap densities were calculated as $5.22\times10^{15}$ $cm^{-3}$ for the control, $3.81\times10^{15}$ $cm^{-3}$ for RIS, $3.27\times10^{15}$ $cm^{-3}$ for BTN, and $5.40\times10^{15}$ $cm^{-3}$ for NSC. These results confirm that RIS and BTN effectively suppress trap density in the perovskite films, enhancing charge extraction and minimizing recombination losses, which are essential for improved PCE. Conversely, NSC-treated films exhibited higher trap densities, consistent with its detrimental impact on device performance.

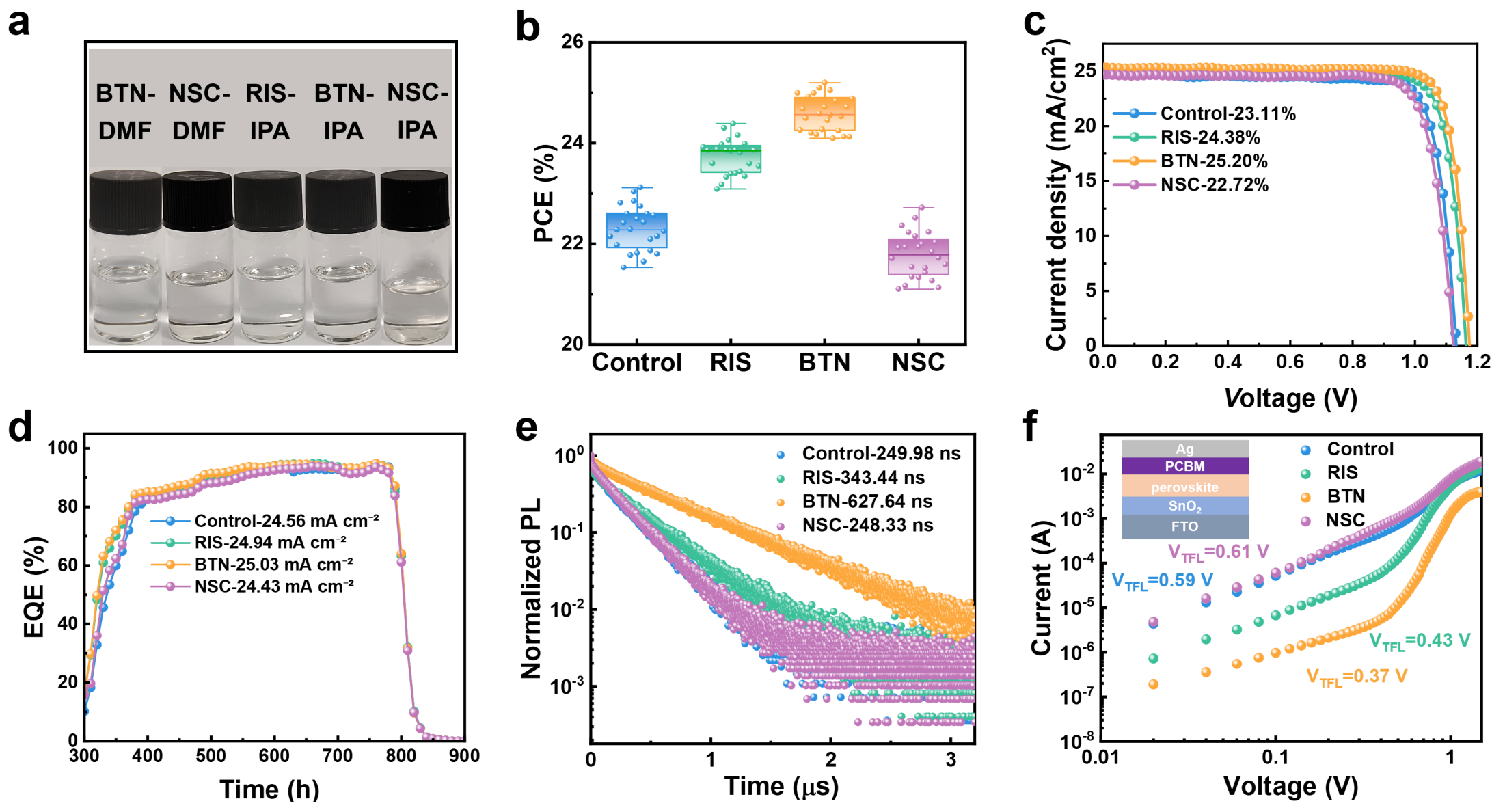

**Figure 5. (a)** Solubility of candidate additives (RIS, BTN and NSC) in DMF and IPA solvents. **(b)** Box plot displaying the distribution of PCEs for control and additive-treated PSCs. **(c)** Current density–voltage (*J-V*) curves for the champion devices. **(d)** EQE curves, **(e)** TRPL spectra and **(f)** SCLC measurements for control and additive-treated PSCs.

## 3. Conclusion

Machine learning (ML) offers significant potential for screening effective organic molecular additives for enhanced device PCE performance; however, predictive biases for novel materials, stemming from limited datasets and reliance on predefined descriptors, limit its effectiveness. The vast chemical space, teeming with millions of molecules exhibiting immense structural diversity, further exacerbates the challenge, making the identification of potential candidates even more intricate. We have therefore developed Co-Pilot for Perovskite Additive Screening (Co-PAS), a robust ML-driven framework that integrates molecular scaffold classification and molecular structure representation via JTVAE-latent vectors for enhanced accuracy in PCE predictions. Co-PAS demonstrated its effectiveness through a domain knowledge-guided, multi-stage screening process. This framework demonstrated the efficient evaluation of an extensive dataset of 250,000 molecules from the PubChem database, which involved verification of JTVAE processibility, prediction of high-efficiency PCE molecules, evaluation of molecular properties related to PCE, and CAS code availability. This adaptable workflow offers flexibility and can be tailored to various research objectives and datasets. Through this workflow, we identified a series of promising additives, with Risedronic acid (RIS) and Boc-L-threonine N-hydroxysuccinimide ester (BTN) achieving experimental PCE values of 24.38% and 25.20%, respectively. The successful experimental validation of these additives highlights the potential of Co-PAS in expediting the discovery of high-performance additives for PSCs. This work significantly advances the application of ML in materials science, paving the way for more effective screening of additives and contributing to the development of high-performance PSCs.

## Acknowledgements

The authors acknowledge fruitful discussions with Prof. Zhen Li from Northwestern Polytechnical University, Prof. Bo Chen from Xi'an Jiaotong University, and Assoc. Prof. Lin Zhu from Nanjing Tech University.

## Author contributions

Y.P. and Z.D. contributed equally to this work. Conceptualization: Z. L., H.-Q. W. Coding: Y.P., Y.Z., N.J., Z. L. Experiments: Z.D., Y. M. Manuscript writing: Y.P., Z. L, R.C. with help and revision feedback from H.-Q. W., H.-Y. W, Y. M. Project Coordination: Z. L, H.-Q. W., R.C.

## Code & data availability

All code and data used to run Co-PAS is available in a GitHub repository at github.com/MatAILab-NPU/Co-PAS. Other data that support the findings of this work are available from the corresponding author upon reasonable request.

## Declaration of interests

The authors declare no competing interests.

# Supplementary Information

# Machine Learning Co-pilot for Screening of Organic Molecular Additives for Perovskite Solar Cells

Yang Pu[1,#], Zhiyuan Dai[1,#], Yifan Zhou[1], Ning Jia[1], Hongyue Wang[1], Yerzhan Mukhametkarimov[2], Ruihao Chen[1*], Hongqiang Wang[1,*], Zhe Liu[1,*]

[1]School of Materials Science and Engineering, State Key Laboratory of Solidification Processing, Northwestern Polytechnical University, 127 Youyi West Road, Xi'an, Shaanxi, P.R. China 710072

[2] Department of Physics and Technology, Al-Farabi Kazakh National University, 71 Al-Farabi Avenue, Almaty, Republic of Kazakhstan, 050040

[#]These authors contributed equally to this work.

[*]Correspondence authors. E-mails: zhe.liu@nwpu.edu.cn (Z. Liu); hongqiang.wang@nwpu.edu.cn (H.-Q. Wang); rhchen@nwpu.edu.cn (R. Chen).

## Experimental Section

### Materials

Fluorine doped tin oxide coated (FTO) glasses (14Ω/sq) were purchased from Advanced Election Technology Co., Ltd. Methylammonium iodide (MAI, 99.5%), formamidinium iodide (FAI, 99.5%), cesium iodide (CsI, 99.5%), and methylammonium chloride (MACl, 99.5%) were purchased from Xi'an Yuri Solar Co., Ltd. Lead iodide ($PbI_2$, >99.99%) and (2-(3,6-Dimethoxy-9H-carbazol-9-yl)ethyl)phosphonic acid (MeO-2PACz, 99.5%) were purchased from TCI America. [6,6]-phenyl-C61-butyric acid methyl ester (PCBM) and 2,9-dimethyl-4,7-diphenyl-1,10-phenanthroline (BCP) were purchased from Vizuchem Co., Ltd. Dimethylformamide (DMF, 99.8%), dimethyl sulfoxide (DMSO, 99.8%), ethyl acetate (EA, 99.8%), isopropanol (IPA, 99.8%) chlorobenzene (CB, 99.5%) and 2-Methoxyethanol (2-Me, 99.8%) were purchased from Sigma-Aldrich. Risedronic acid (RIS), Boc-L-threonine N-hydroxysuccinimide ester (BTN) were purchased from Shanghai Macklin Biochemical Technology Co., Ltd. and NSC-2805 was purchased from TargetMol Chemicals Inc. All the chemicals are used directly without further purification.

## Device fabrication

FTO glass substrate was ultrasonically cleaned by detergent, deionized water, acetone and ethanol for 25 min, respectively. The substrate was dried by nitrogen and activated by plasma for 20 min, then transferred into a nitrogen glove box. The hole transport layer (HTL) was fabricated by spin-coating the MeO-2PACz (0.7 mg/mL in 2-Me) on FTO substrate at 3000 rpm for 30 s (spin coater from Jiangyin J. Wanjia Technology Co., Ltd.), followed by annealing at 100 °C for 10 min. The 1.5 M perovskite precursor solution with the composition of $FA_{0.85}MA_{0.10}Cs_{0.05}PbI_3$ was prepared by dissolving MAI, FAI, CsI, $PbI_2$ and excess MACl (5 mol%) in a mixed solvent of DMF and DMSO with a volume ratio of 4:1. For bulk passivation, the perovskite precursors were obtained by adding different molecules (BTN: 0.5% molar ratio, NSC-2805: 0.5% molar ratio) into precursor. 50 μL of perovskite precursor was dropped on HTL and spin-coating at 5000 rpm for 30 s, 200 μL EA was dropped on the film at 5 s before the end of procedure. The film was then annealed at 120 °C for 30 min. For interface passivation, after cooling down, 40 μL of different molecule solutions (RIS: 0.3 mg/mL, BTN: 0.5 mg/mL, NSC-2805: 0.5 mg/mL in IPA) were dynamically spin-coated on the films at 4000 rpm for 30 s and annealed at 100 °C for 5 min to obtain additive-treated samples. The electron transport layer (ETL) was fabricated by spin-coating the PCBM (20 mg/mL in CB) on films at 2000 rpm for 30 s, and 150 μL BCP (0.75 mg/ mL in IPA) was dynamically spin-coated on ETL at 4000 rpm for 30 s, followed by thermal annealing at 70 °C for 5 min. Finally, 100 nm thickness of Ag was thermally evaporated at a rate of 0.4 Å/s to complete the solar cell devices. The active area is 0.05 $cm^2$ by black mask.

## Characterization

The time-resolved photoluminescence (TRPL) spectra were obtained by a Pico Quant Fluo Time 300 fluorescence spectrometer. The external quantum efficiency (EQE) of devices were measured by a Enlitech EQE measurement system (QE-R3011). *J-V* curves characteristics and the space charge limited current (SCLC) of PSCs were measured using solar simulator (Oriel 67005, 150 W) with irradiation intensity of 100 mW $cm^{-2}$ AM 1.5 G standard light.

# Machine Learning Section

## Molecular Scaffold Classifier (MSC)

Using RDKit, the scaffold of molecule can be generated using the Murcko scaffold tool, which focus on extracting the core structure by removing peripheral side chains and retaining the main framework[1][2]. The scaffold typically consists of the fused ring system and any connecting linkers, representing the structural backbone that is common to a family of compounds. The key advantage of this approach lies in its ability to simplify molecular representations, allowing for the efficient extraction of structural information. In this work, we further divided the molecules in the dataset into nine major categories based on their Murcko scaffolds. This classification enhanced the model's ability to efficiently extract relevant molecular information. The nine molecular scaffold categories are shown in Figure S1.

Following the division of additive molecules into these nine scaffold-based categories, stratified sampling was implemented to ensure a representative distribution of each category within the dataset. Given the variation in molecular counts across these groups, different sampling strategies were applied for constructing the training and testing sets. Specifically, for groups with a smaller sample size (containing 3 or fewer molecules), 1 molecule was randomly selected as the test set, while the remainder constituted the training set. This approach maintains model training and evaluation efficacy even under limited sample conditions. Conversely, in groups with larger sample sizes (containing more than 3 molecules), 2 molecules were randomly designated as test samples, with the rest forming the training set. This sampling method leverages the additional data to enhance model generalization. Using this stratified sampling approach, the entire dataset was ultimately split into training and test sets with a 9:1 ratio.

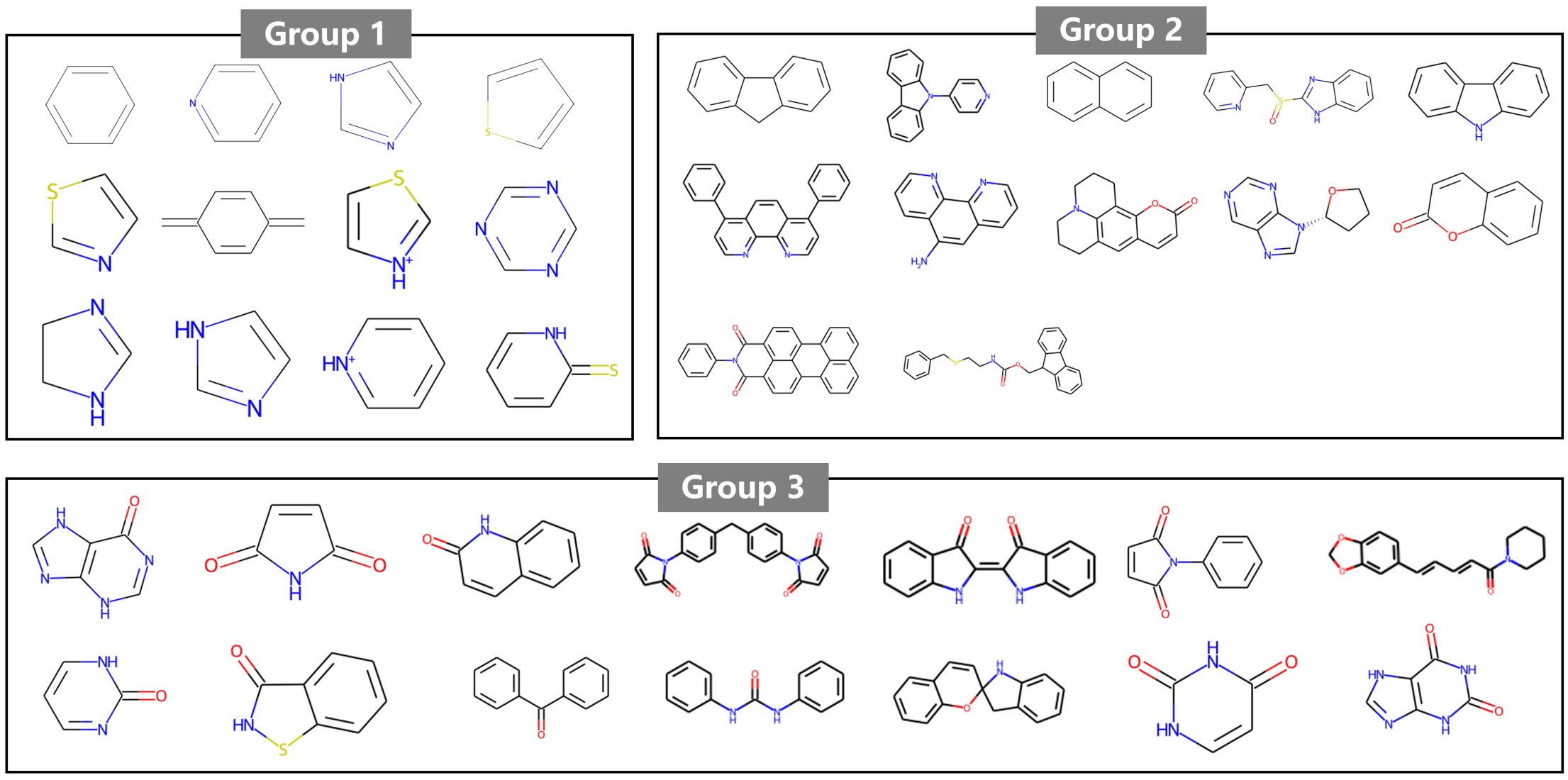

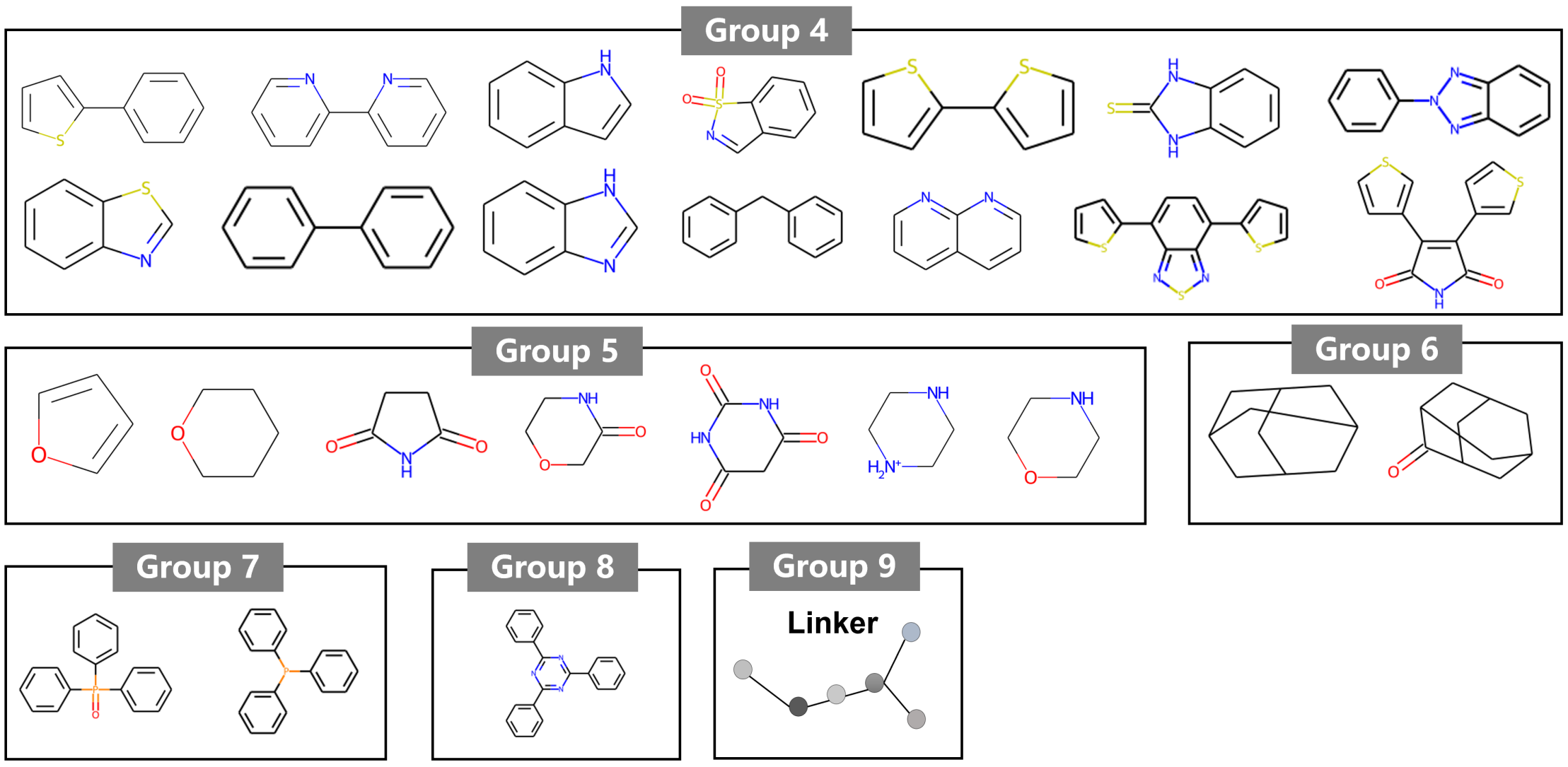

Figure S1. Representation of the nine molecular scaffold categories.

In order to demonstrate the robustness of MSC, we conducted a thorough comparison with the traditional random splitting approach and Leave-One-Group-Out (LOGO) strategy. Due to the relatively small dataset, we ran the model training process 200 times and calculated the average and standard deviation of the resulting metrics. To achieve accuracy and correlation reliability, we employed the Mean Absolute Error (MAE) and Spearman coefficient as evaluation metrics. Table S1 presents the average MAE and Spearman Coefficient along with their standard deviations for MSC, Random and LOGO method with various model inputs.

Table S1. Average MAE and Spearman Coefficient with Standard Deviation for MSC, Random and LOGO method with different model Inputs.

| Method | MACCS | | RDKit Descriptors | | MACCS+ RDKit Descriptors | |
|---|---|---|---|---|---|---|
| | MAE | Spearman | MAE | Spearman | MAE | Spearman |
| MSC | 1.1327 ± 0.2708 | 0.5982 ± 0.2152 | 0.9854 ± 0.2729 | 0.6811 ± 0.2049 | 0.9696 ± 0.2471 | 0.6670 ± 0.2048 |
| Random | 1.9278 ± 0.4423 | 0.1463 ± 0.2712 | 1.9635 ± 0.4503 | 0.2009 ± 0.2668 | 1.9853 ± 0.4092 | 0.1322 ± 0.2904 |
| LOGO | 2.2101 ± 0.6195 | 0.0868 ± 0.4183 | 1.7329 ± 0.4684 | 0.0600 ± 0.6272 | 1.81261 ± 0.4874 | 0.0241 ± 0.6071 |

## Dataset for ML training

The dataset was compiled from 129 realistic additive molecules used in perovskite solar cells (PSCs), sourced from published literature (refer to the EXCEL file: Supporting Information-Molecules Data). These molecules were converted into canonical SMILES strings[3] and then transferred to molecular fingerprints descriptors and JTVAE-latent vectors. The generation of Molecular Access System (MACCS) and descriptors used the RDKit toolkit[4]. JTVAE-latent vectors are derived from the Junction Tree Variational Autoencoder (JTVAE)[5].

All the molecular descriptors were refined. Initially, 208 descriptors were generated using RDKit. Descriptors with a constant value of zero across all samples were removed, leaving 184 descriptors. The remaining descriptors were then normalized using max normalization. Max normalization scales data values to a range of [0, 1], using the maximum value in the data set as the reference point. The formula for max normalization is as follows:

$$X' = \frac{X}{X_{\max}}$$

Where $X'$ is the normalized value after scaling, $X$ is the original value of the descriptor, $X_{\max}$ is the maximum value of the descriptor across the dataset. This method ensures that all values are proportionally scaled between 0 and 1, with the maximum value being 1. Next, we used the Variance Threshold for feature selection of the remaining 184 descriptors[6]. Standard deviation (STD) is a measure of the amount of variation or dispersion of a set of values. It indicates how much the individual data points deviate from the mean of the data set. The formula is as follows:

$$\mathrm{s} = \sqrt{\frac{1}{n-1}\sum_{i=1}^{n}(x_i - \bar{x})^2}$$

Where: s is the sample standard deviation, n is the number of data points (sample size), $x_i$ is the value of the i-th data point, $\bar{x}$ is the mean (average) of the data set. Descriptors with a low variance (STD value), defined by a threshold of 0.2, were excluded, reducing the set to 49 descriptors. Figure S2 and S3 show the distributions of STD values for descriptors before and after applying the Variance Threshold.

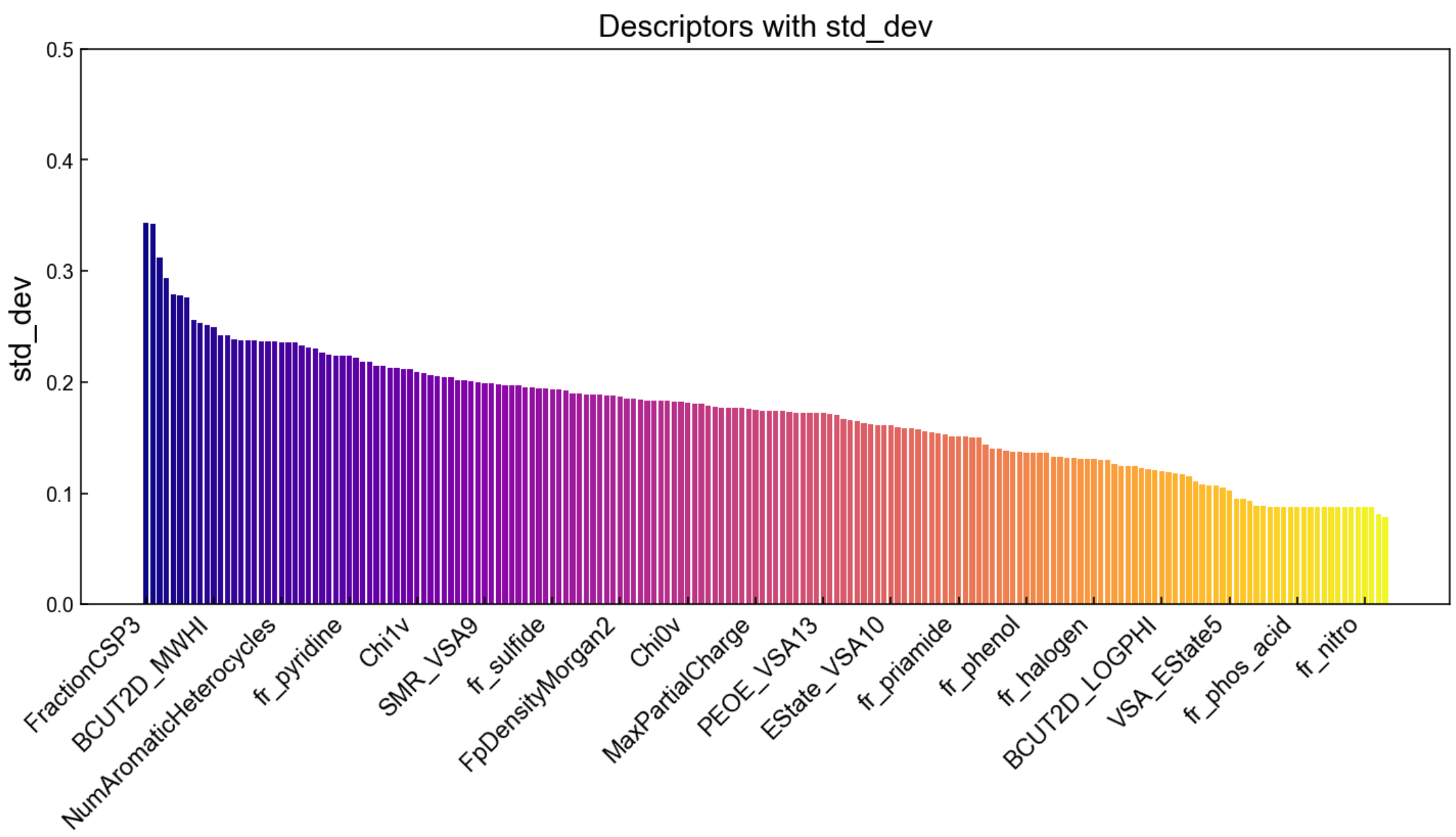

Figure S2. The STD distribution of 184 descriptors before Variance Threshold processing.

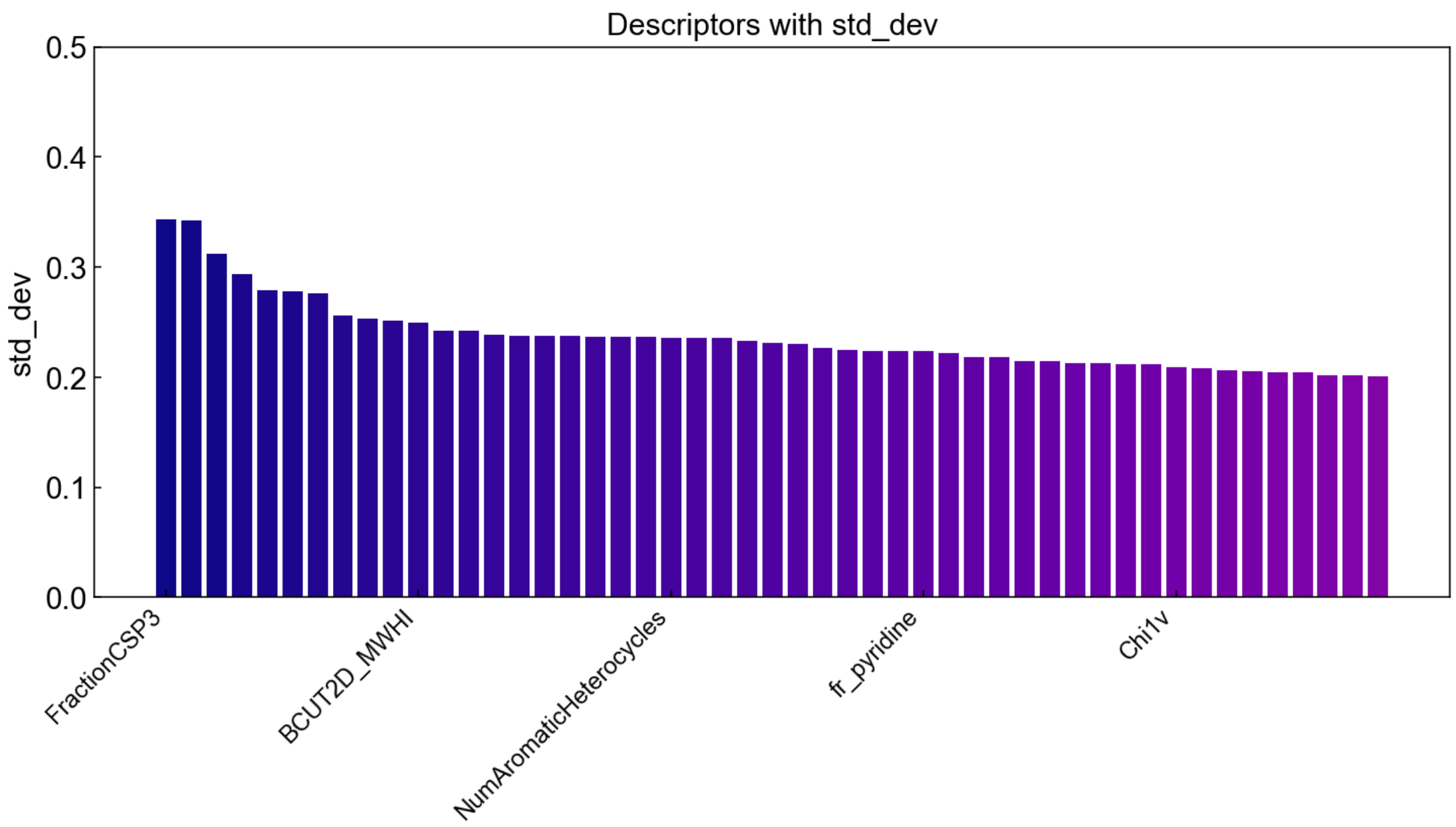

Figure S3. The STD distribution of 49 descriptors after Variance Threshold processing.

Next, Pearson correlation coefficients (PCCs) were calculated for each pair of the remaining descriptors[7]. Descriptors with an absolute PCC greater than 0.9 were grouped together, and within each group, only one descriptor was retained. After this filtering, 5 descriptors were removed, leaving 44 descriptors, which were used for model training. The PCC is defined as:

$$\rho_{X,Y} = \frac{\sum_{i=1}^{n}(x_i - \overline{x})(y_i - \overline{y})}{\sqrt{\sum_{i=1}^{n}(x_i - \overline{x})^2}\sqrt{\sum_{i=1}^{n}(y_i - \overline{y})^2}}$$

Where $x_i$ and $y_i$ represent individual values of two descriptors, $\overline{x}$ and $\overline{y}$ are the means of the descriptors, and n is the number of molecules. The rest of 44 descriptors were performed normalization using the MinMaxScaler function when inputting them into models. The Correlation matrix of Pearson coefficients between 44 selected descriptors is shown in Figure S4.

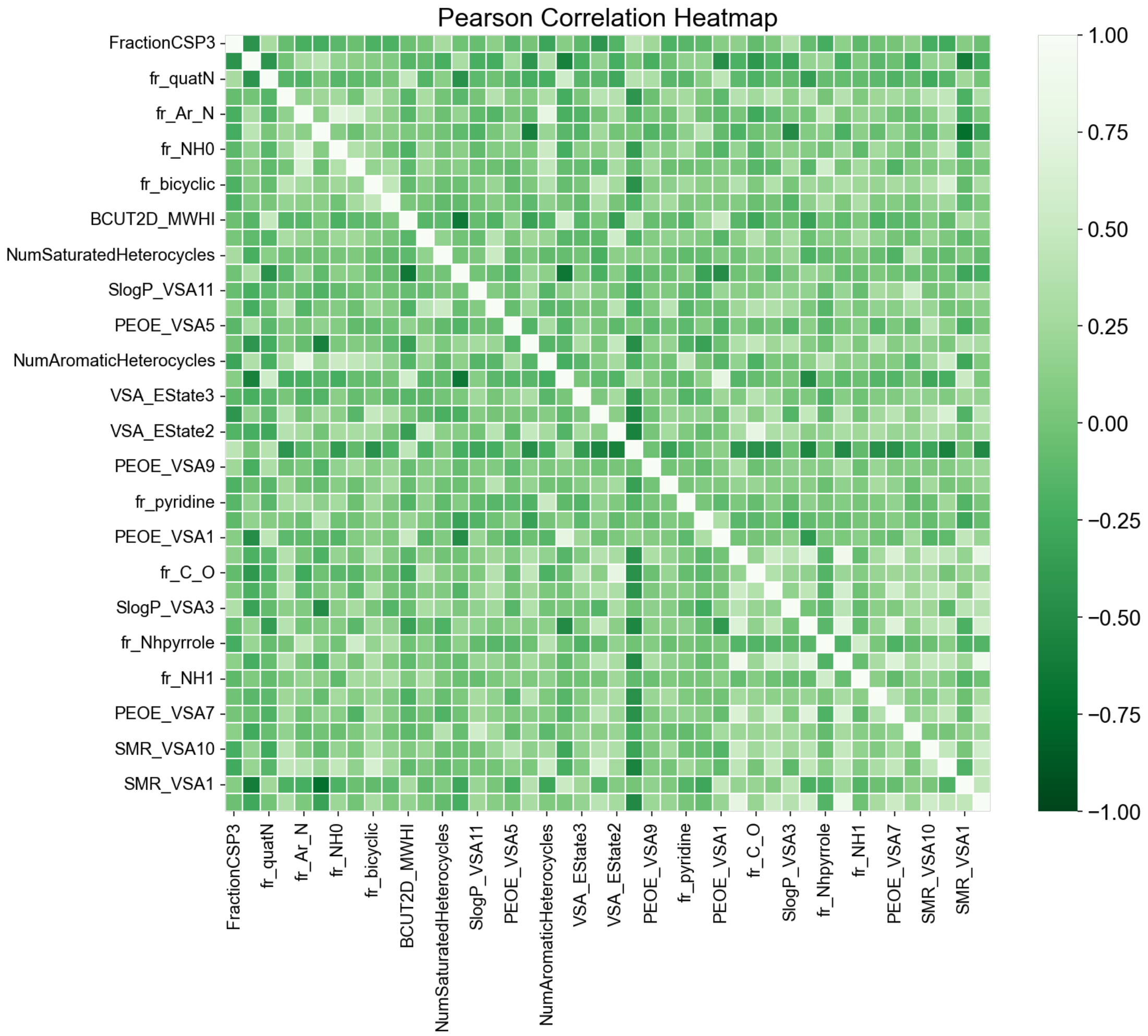

Figure S4. Correlation matrix of Pearson coefficients between 44 selected descriptors.

## JTVAE-Latent Vectors

The Junction Tree Variational Autoencoder (JTVAE) stands out as an innovative framework designed specifically for molecular graph generation. Its distinctive architecture allows for the effective capture of the complex features inherent in molecular structures. By utilizing both the Junction Tree representation and Graph representation, JTVAE adopts a dual approach that facilitates the learning of comprehensive molecular features[5].

### Representation of Molecular Features in the Latent Space

In JTVAE, the latent space integrates both the hierarchical structure from the Junction Tree Encoder and the connectivity patterns from the Graph Encoder, effectively capturing the distinct features of the molecule “OC(=O)c1csc(Cl)n1” (2-chloro-1,3-thiazole-4-carboxylic acid). This compressed latent representation encodes critical details: the carboxylic acid group branching off the thiazole ring, the placement of the chlorine atom, and the cyclic nature of the thiazole core. By combining these hierarchical and topological elements, the latent space allows JTVAE to reconstruct or generate diverse, structurally valid molecules with similar functional groups, ensuring that essential features like ring structure, branching, and functional group placement are preserved. Figures S5–7 illustrate the two-dimensional latent space distributions for the overall molecule, the junction tree of molecule, and the molecular graph representations, respectively.

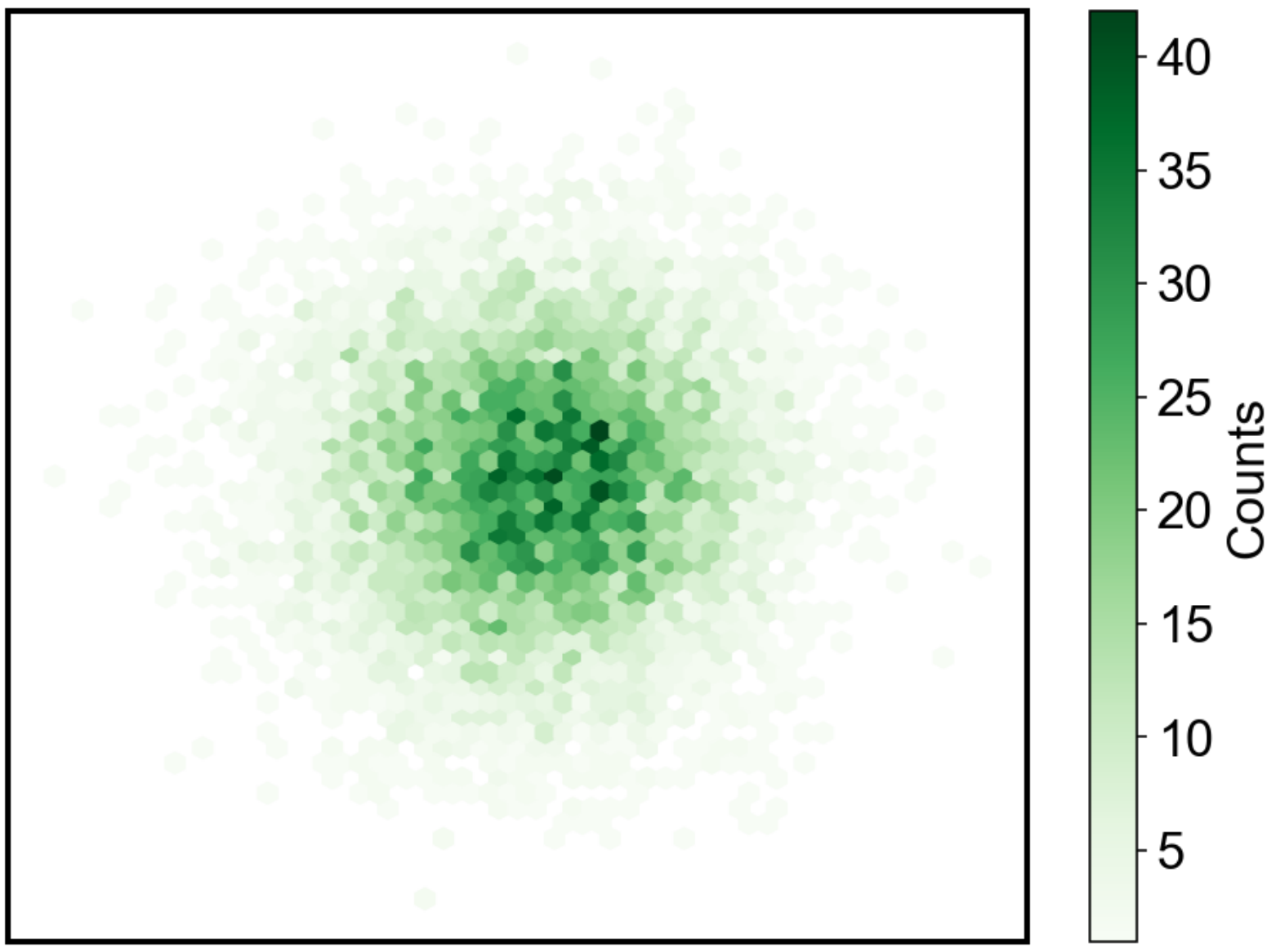

Figure S5. Two-dimensional representation of latent space for JTVAE.

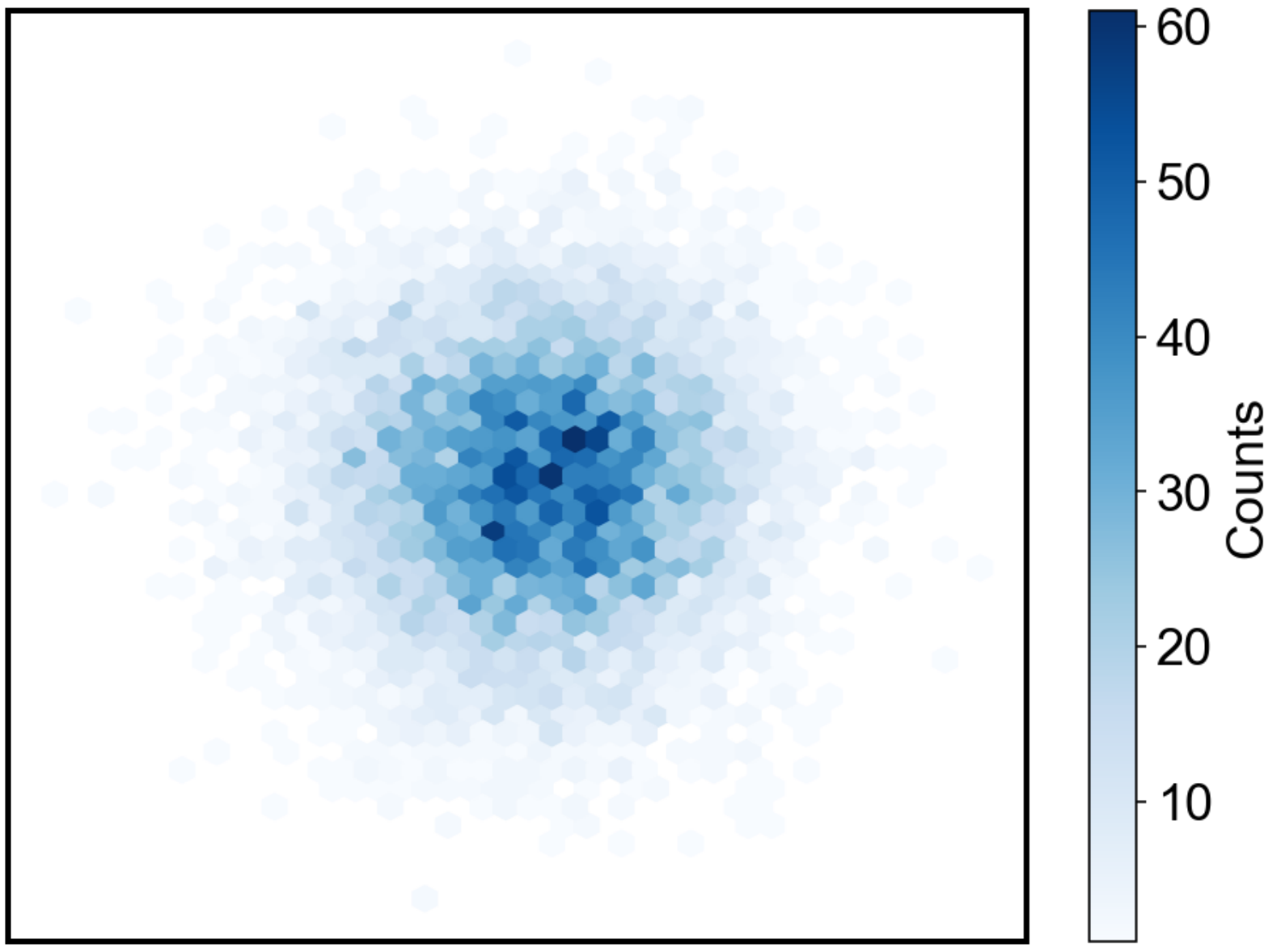

Figure S6. Two-dimensional representation of latent space for junction tree autoencoder.

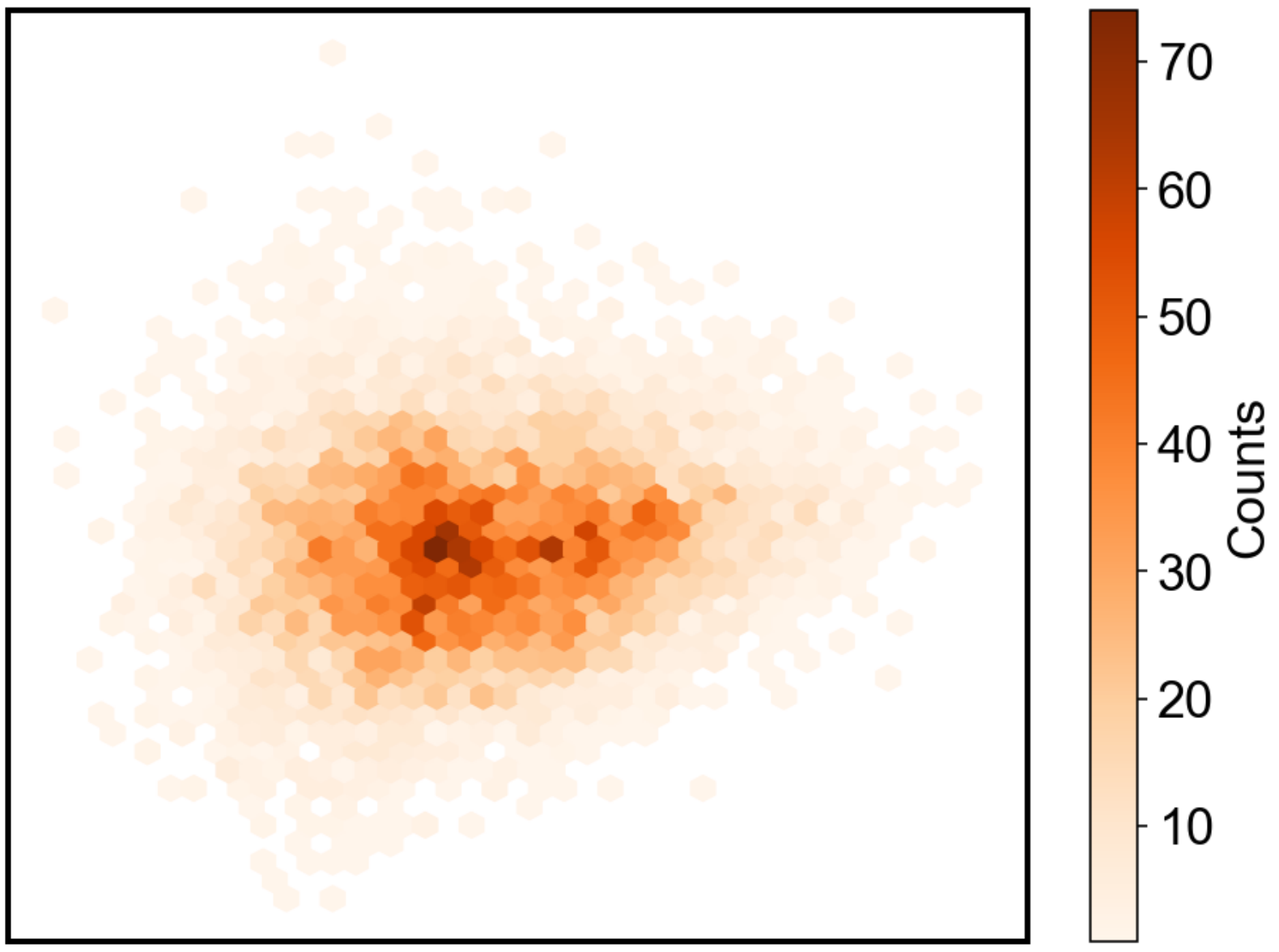

Figure S7. Two-dimensional representation of latent space for graph autoencoder.

**Junction Tree Encoder**

The Junction Tree Encoder effectively captures the hierarchical structure of molecules using a junction tree representation, allowing for the decomposition of complex molecular graphs into simpler components or "fragments." For example, consider the molecule represented by the SMILES notation “OC(=O)c1csc(Cl)n1”, which corresponds to 2-chloro-1,3-thiazole-4-carboxylic acid. In the junction tree representation, nodes correspond to substructures such as the carboxylic acid group (-COOH), and the thiazole ring, while the edges denote the relationships between these substructures, indicating how the carboxylic acid is attached to the aromatic ring (Figure S8). This representation captures crucial features, including functional groups vital for the molecule's reactivity, and connectivity patterns that reflect interactions among functional groups, such as the influence of the chlorine atom in the aromatic ring on electronic properties.

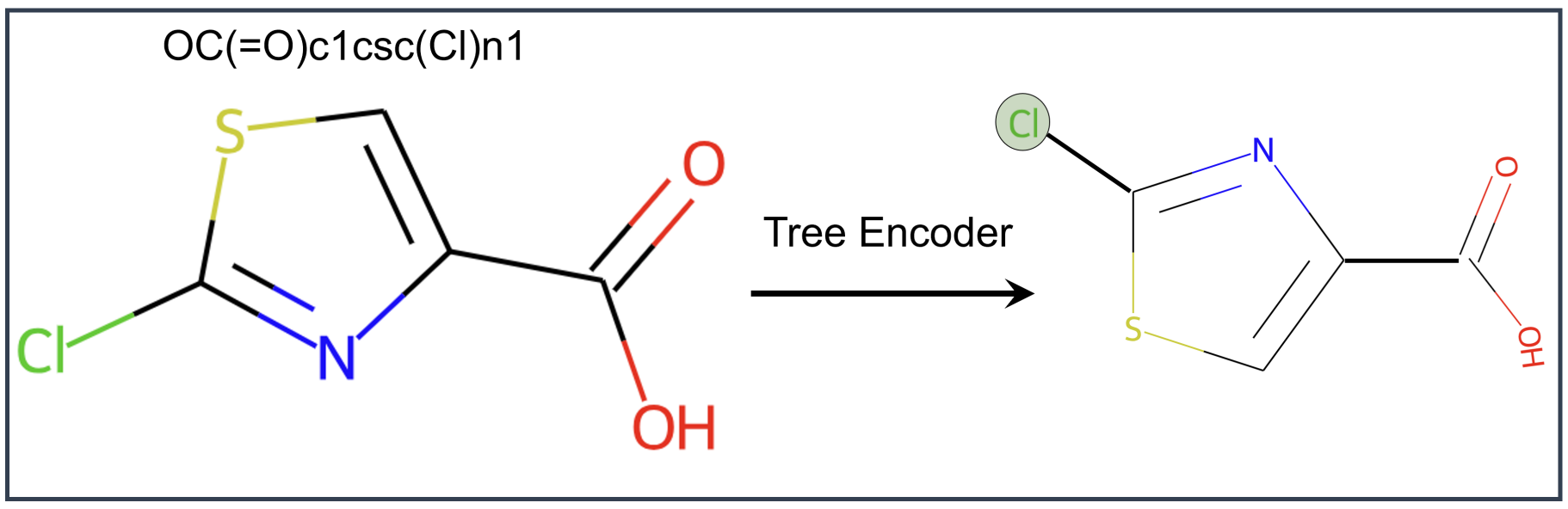

Figure S8. Illustration of junction tree encoder to capture molecule information including scaffold and functional groups.

## Graph Encoder

Complementing the Junction Tree Encoder, the Graph Encoder captures the connectivity and relationships among atoms in a molecular graph. It focuses on the overall topology of the molecule, allowing the model to recognize patterns and dependencies that arise from the arrangement of atoms and bonds. The Graph Encoder facilitates the learning of complex relationships, such as ring structures and branching, enabling it to encapsulate properties that influence molecular behavior. For the molecule mentioned above (2-chloro-1,3-thiazole-4-carboxylic acid), the Graph Encoder represents each atom as a node, with edges denoting the bonds between them. For example, carbon atoms, a nitrogen atom, a sulfur atom, and a chlorine atom are nodes, while single and double bonds between these atoms are captured as edges. As shown in Figure S9, in the generated molecular visualization of 2-chloro-1,3-thiazole-4-carboxylic acid, the numbers 1-8 correspond to specific atoms or bonds highlighted in the structure. This indexing aids in visualizing the relationships and connectivity between different functional groups, such as the carboxylic acid group and the thiazole ring, enhancing the interpretation of the molecular properties and behavior.

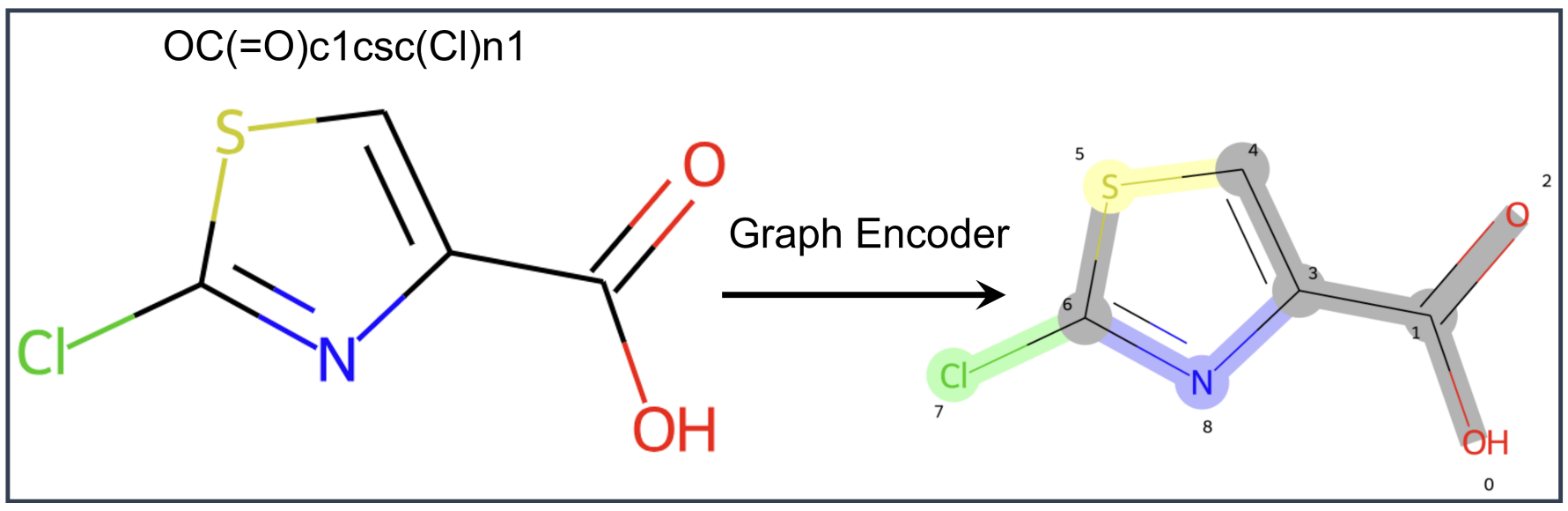

Figure S9. Illustration of graph encoder to capture molecule information including atoms and bond type.

**Neighborhood Visualization**

We construct a grid visualization to explore the molecular neighborhood of "OC(=O)c1csc(Cl)n1" (2-chloro-1,3-thiazole-4-carboxylic acid) The molecule is initially encoded into the latent space, with two orthogonal unit vectors randomly generated to serve as the grid axes. By shifting along combinations of these vectors, we obtain a set of latent vectors, which are then decoded back into their corresponding molecules. Figure S10 illustrates the resulting local neighborhood for this molecule.

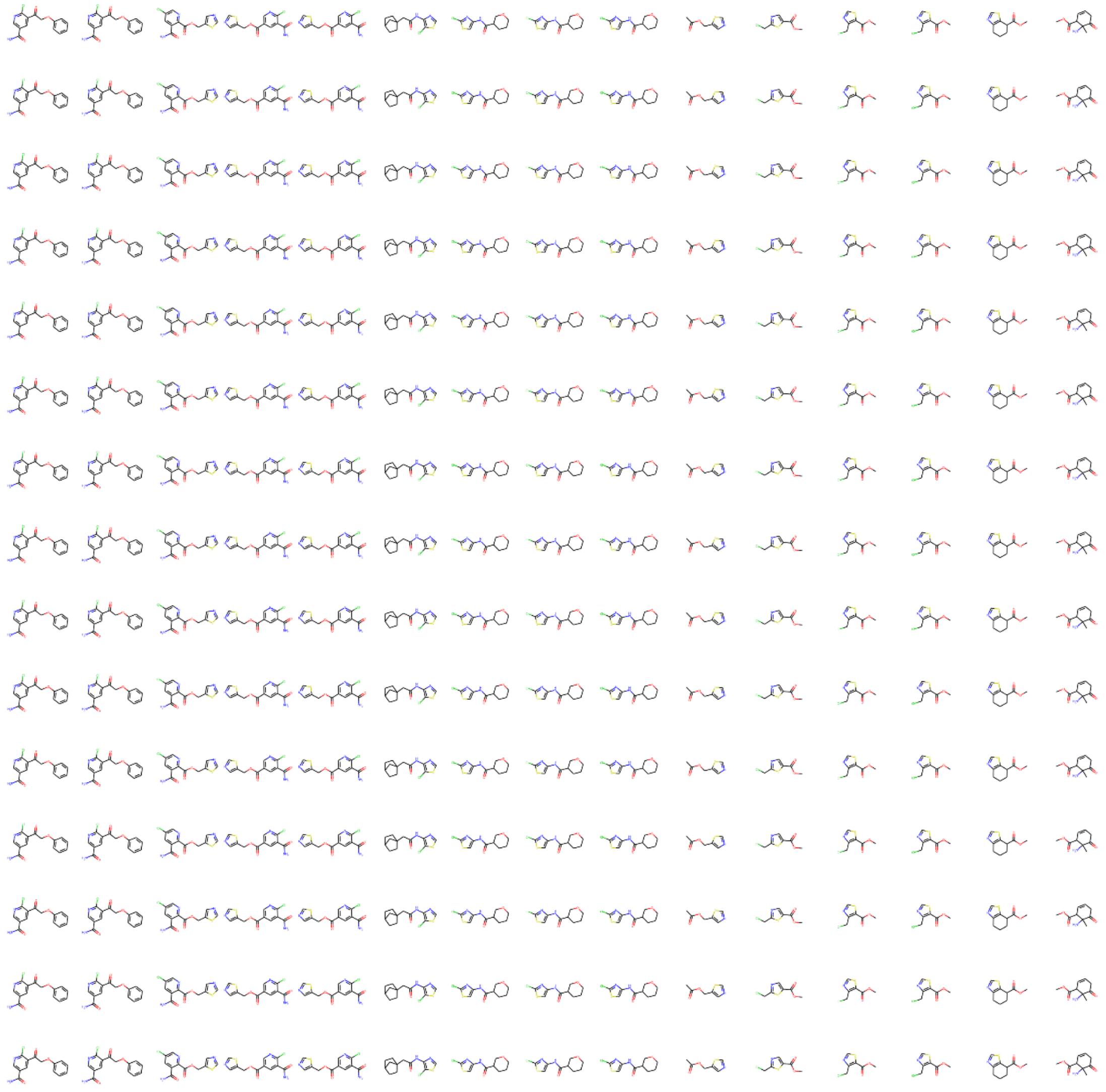

Figure S10. Neighborhood visualization of molecule OC(=O)c1csc(Cl)n1.

## Model Optimization and Evaluation

In order to demonstrate the effectiveness of incorporation of JTVAE-latent vectors, we selected several popular regression algorithms, including Random Forest (RF) regression, Gradient Boosting (GB) regression, and Support Vector Machine (SVM) regression. Their optimized hyperparameters of these models are shown in Table S2. Table S3 presents the average MAE and the corresponding standard deviations for RF,

GB, and SVR across various model inputs.

Table S2. The hyperparameters of RF, GB, and SVR regressors.

| **Regressor** | **Hyperparameters** |
|---|---|
| RF | n_estimators = 55; max_depth = 10 |
| GB | n_estimators = 35; max_depth = 4 |
| SVR | c = 500; epsilon = 0.75 |

Table S3. Average MAE with Standard Deviation for model RF, GB, and SVR with different model Inputs.

| Model \ Input | M | D | M+D | Z | Z+M | Z+D | Z+M+D |
|---|---|---|---|---|---|---|---|
| RF | 1.1349 ± 0.2718 | 0.9725 ± 0.2402 | 0.9781 ± 0.2763 | 1.0528 ± 0.2507 | 1.0355 ± 0.2584 | 0.9972 ± 0.2356 | 1.0073 ± 0.2162 |
| GB | 1.1327 ± 0.2708 | 0.9854 ± 0.2729 | 0.9696 ± 0.2471 | 1.0047 ± 0.2474 | 0.9890 ± 0.2646 | 0.8298 ± 0.2477 | 0.8965 ± 0.2631 |
| SVR | 1.0578 ± 0.2267 | 1.2143 ± 0.2702 | 1.0732 ± 0.2260 | 1.0078 ± 0.2854 | 0.9590 ± 0.2717 | 1.1106 ± 0.2642 | 0.9788 ± 0.2528 |

## Validation Dataset

Table S4. Validation dataset from in the published papers.

| SMILES | PCE | Doi |
|---|---|---|
| Nc1ccncc1 | 16.9 | 10.1016/j.cej.2022.137033 |
| CC(N)=O | 22.7 | 10.1016/j.cej.2022.138559 |
| CN(C)CCO | 20.69 | 10.1002/solr.202101107 |
| CC(C)(S)[C@@H](N)C(O)=O | 24.09 | 10.1002/solr.202200567 |
| CCCCCl | 21.35 | 10.1002/solr.202100979 |
| OC(=O)c1cc(cc(c1)C(F)(F)F)C(F)(F)F | 21.09 | 10.1021/acsami.1c18035 |
| NCc1ccccc1Cl | 22.6 | 10.1021/acsami.2c02250 |
| c1ccc(cc1)P(c2ccccc2)c3ccccc3 | 20.22 | 10.1002/cssc.202102189 |
| COc1ccc(CC(O)=O)cc1 | 22.32 | 10.1016/j.jpowsour.2021.230734 |
| SCCS | 21 | 10.1002/eom2.12185 |
| OC(=O)Cc1sccc1 | 20.6 | 10.1039/d2cc03890k |
| OC(=O)Cc1cscc1 | 20 | 10.1039/d2cc03890k |
| NC(=S)C(N)=S | 21.06 | 10.1016/j.mtener.2022.101004 |
| NC(=O)Cc1ccccc1 | 22.98 | 10.1002/aenm.202203635 |
| [NH3+]Cc1cccc2ccccc12 | 19.6 | 10.1016/j.cej.2023.141788 |
| OC(=O)CCC(O)=O | 21.81 | 10.1016/j.cej.2023.141705 |
| NCc1ccc(cc1)C(F)(F)F | 22.36 | 10.1002/aesr.202200128 |
| [O-]C(=O)c1c(F)c(F)c(F)c(F)c1C([O-])=O | 23.7 | 10.1002/solr.202201025 |
| CCOC(=O)C(=C)C#N | 20.12 | 10.1002/solr.202300105 |
| NC(Cc1ccccc1)C(O)=O | 21.95 | 10.1002/smtd.202200669 |
| [K+].[K+].[K+].[O-]C(=O)CN(CC([O-])=O)CC([O-])=O | 21 | 10.1016/j.mtchem.2022.101362 |
| NNC(=O)c1ccccc1 | 22.75 | 10.1016/j.mtener.2023.101269 |
| O[P](O)(=O)c1ccccc1 | 22.02 | 10.1016/j.solmat.2023.112278 |
| Clc1ccccc1Cl | 20.75 | 10.1021/acssuschemeng.2c05816 |

# Candidate Additives Identified by Co-PAS

Table S5. Candidate additives identified by Co-PAS, with relevant compounds from prior literature.

**Screened Molecules**

| Candidate | Scaffold |
|---|---|
| 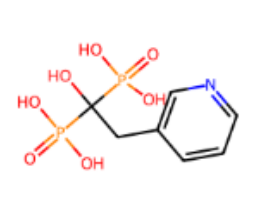 | 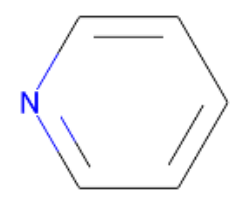 |
| 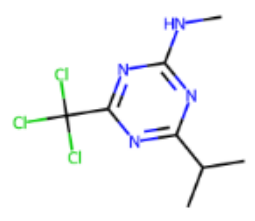 | 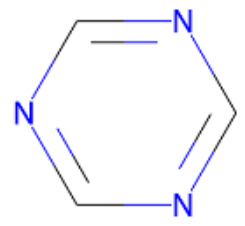 |
| 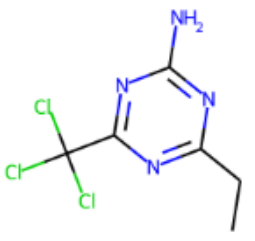 | 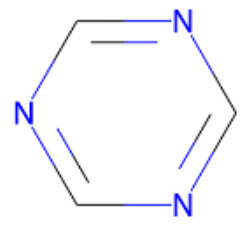 |
| 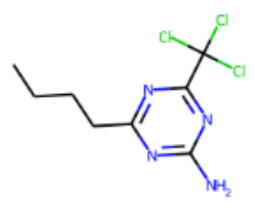 | 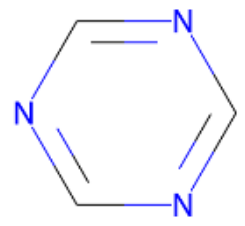 |
| 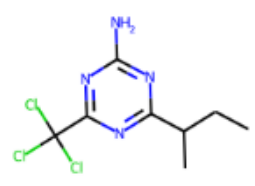 | 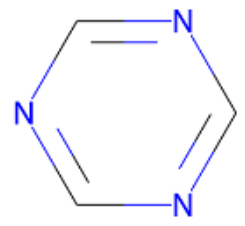 |
| 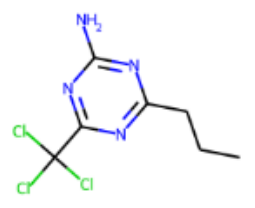 | 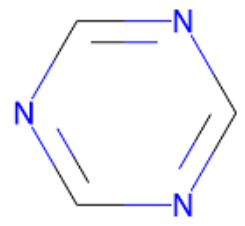 |
| 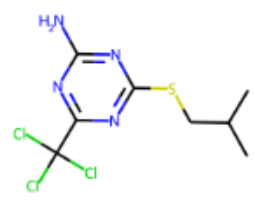 | 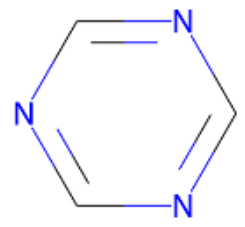 |
| 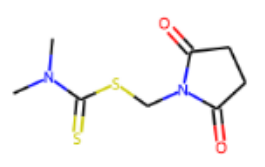 | 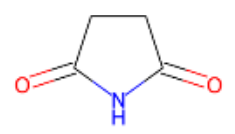 |
| 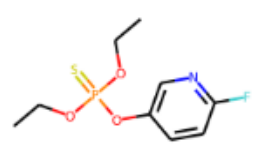 | 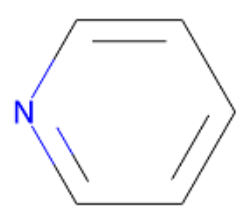 |
| 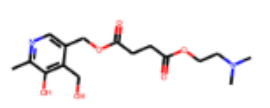 | 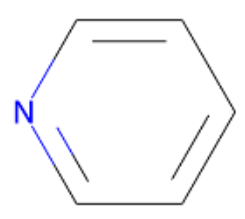 |
| 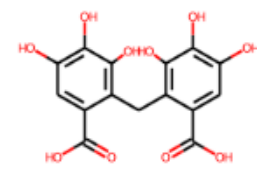 | 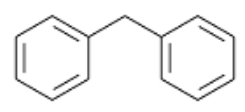 |
| 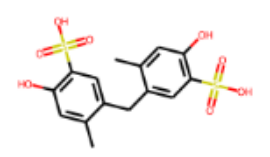 | 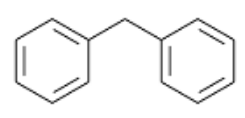 |

**Literature Reported Molecules**

| Reported molecule | PCE | Doi | |
|---|---|---|---|
| 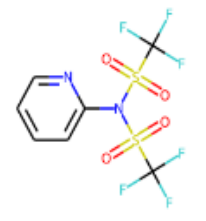 | 21.96% | 10.1016/j.cej.2022.139345 | |
| 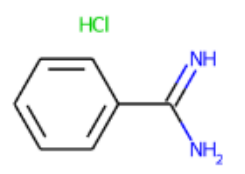 | 22.47% | 10.1016/j.cej.2023.143392 | |
| 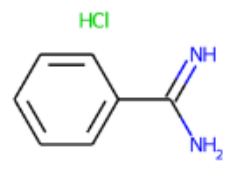 | 22.47% | 10.1016/j.cej.2023.143392 | |
| 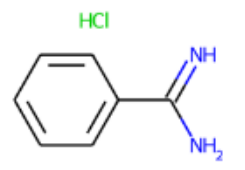 | 22.47% | 10.1016/j.cej.2023.143392 | |
| 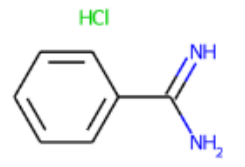 | 22.47% | 10.1016/j.cej.2023.143392 | |
| 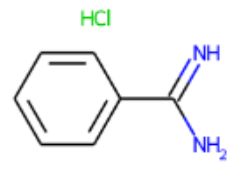 | 22.47% | 10.1016/j.cej.2023.143392 | |
| 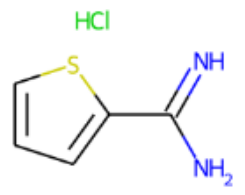 | 23.28% | 10.1016/j.cej.2023.143392 | |
| 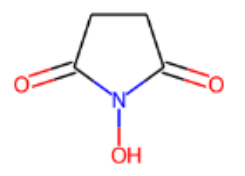 | 22.21% | 10.1002/solr.202200502 | |
| 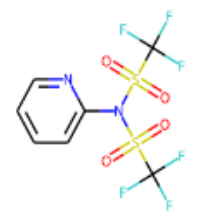 | 21.96% | 10.1016/j.cej.2022.139345 | |
| 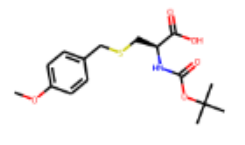 | 21.75% | 10.1002/adma.202301140 | |
| 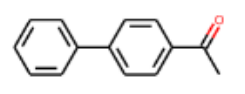 | 22.47% | 10.1002/anie.202302462 | |
| 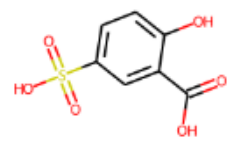 | 21.09% | 10.1039/D1TA07505E | |

| | | | | |
|---|---|---|---|---|
| 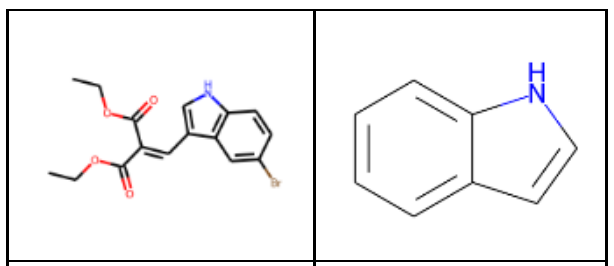 | 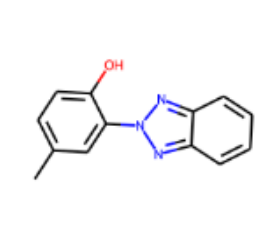 | 22.46% | 10.1016/j.jechem.2021.09.027 | |
| 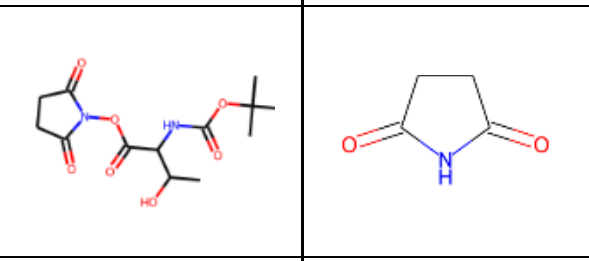 | 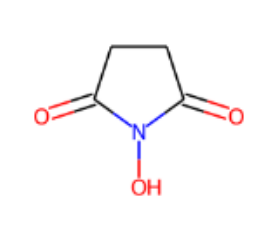 | 22.21% | 10.1002/solr.202200502 | |
| 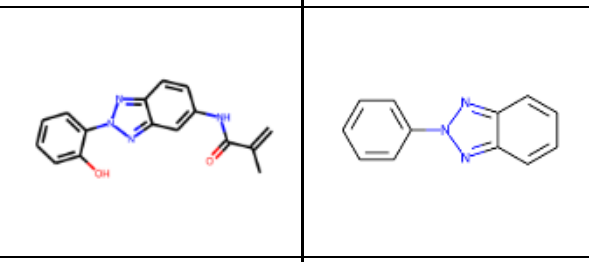 | 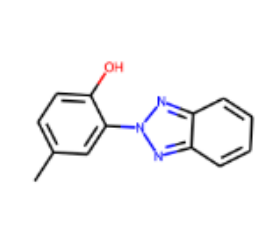 | 22.46% | 10.1016/j.jechem.2021.09.027 | |
| 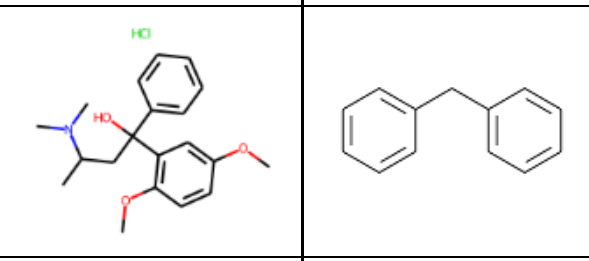 | 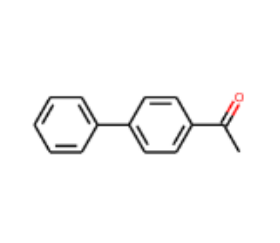 | 22.47% | 10.1002/anie.202302462 | |
| 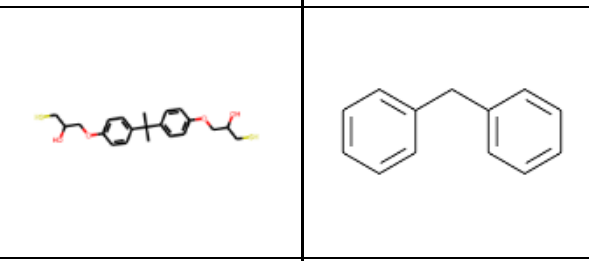 | 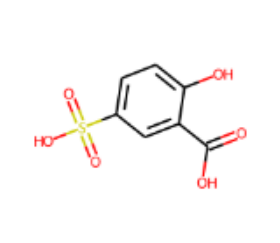 | 21.09% | 10.1039/D1TA07505E | |
| 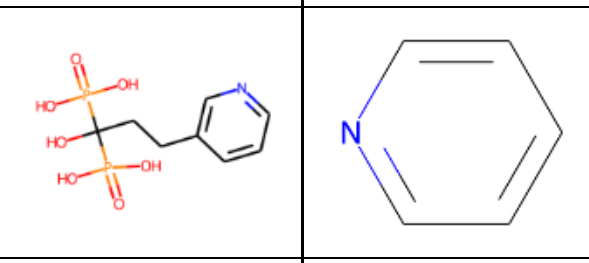 | 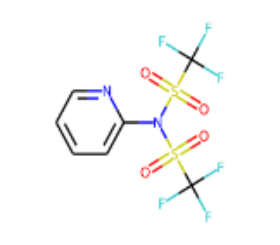 | 21.96% | 10.1016/j.cej.2022.139345 | |
| 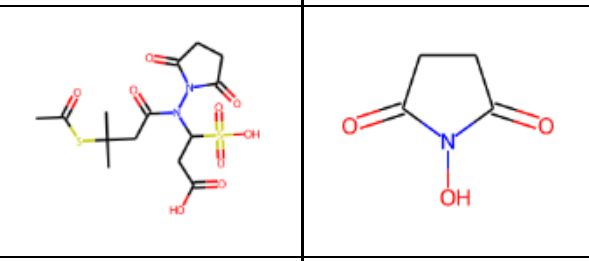 | 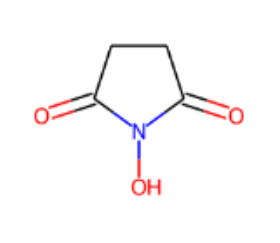 | 22.21% | 10.1002/solr.202200502 | |
| 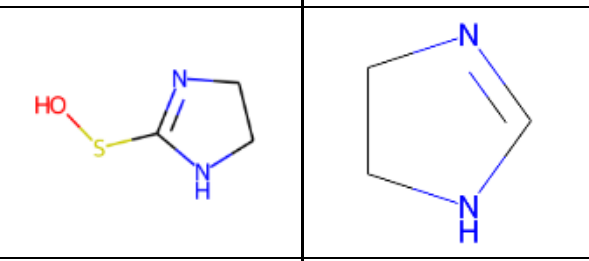 | 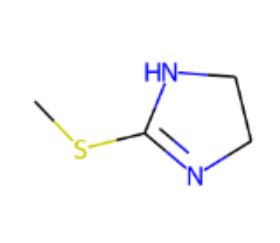 | 21.25% | 10.1021/acsami.1c23991 | |
| 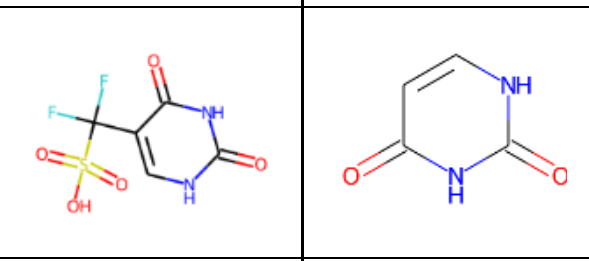 | 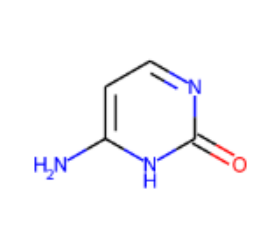 | 22.08% | 10.1016/j.cej.2022.139535 | |
| 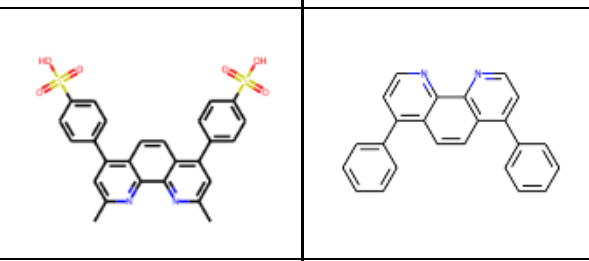 | 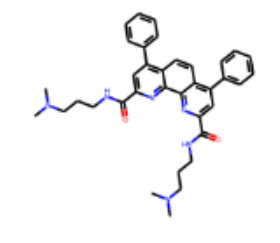 | 20.98% | 10.1002/solr.202200559 | |
| 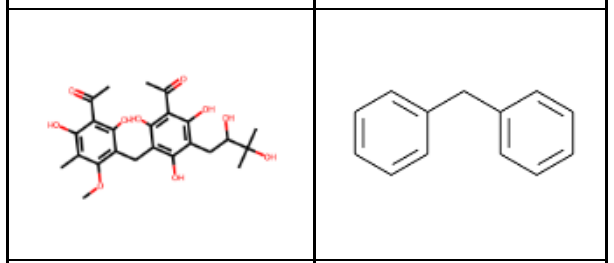 | 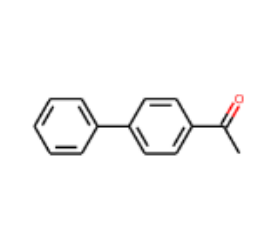 | 22.47% | 10.1002/anie.202302462 | |
| 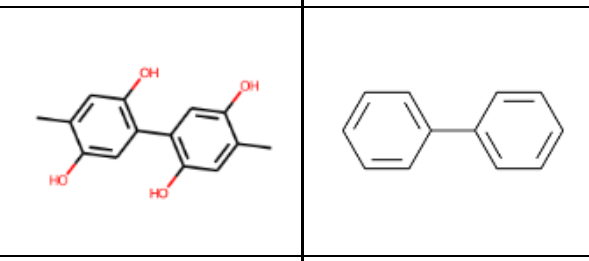 | 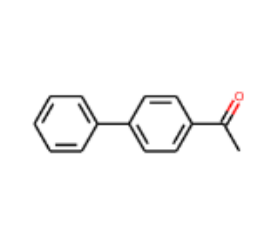 | 22.47% | 10.1002/anie.202302462 | |
| 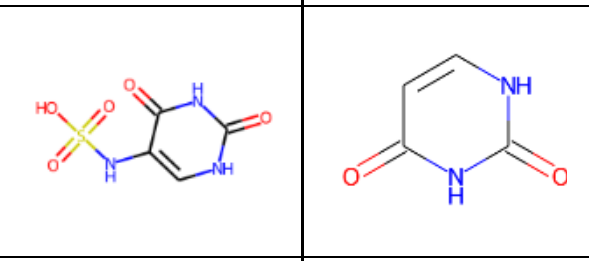 | 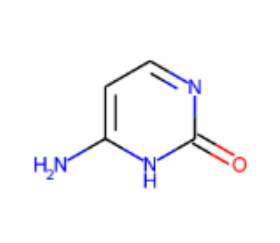 | 22.08% | 10.1016/j.cej.2022.139535 | |
| 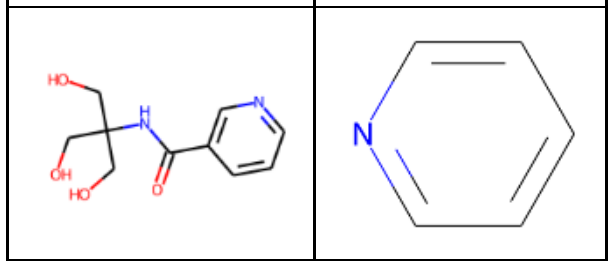 | 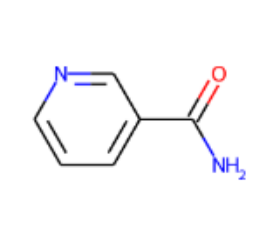 | 23.99% | 10.1021/acsami.0c12030 | |

| | | | | | |
|---|---|---|---|---|---|
| 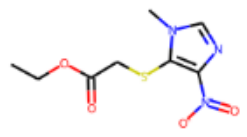 | 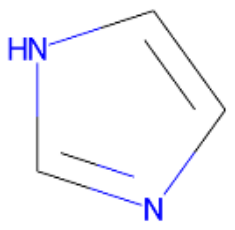 | 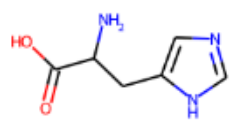 | 22.91% | 10.1021/acsami.2c13585 | |
| 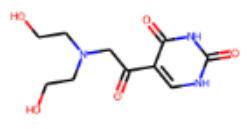 | 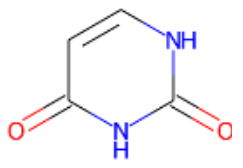 | 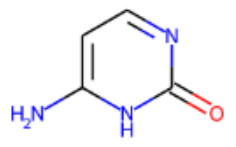 | 22.08% | 10.1016/j.cej.2022.139535 | |
| 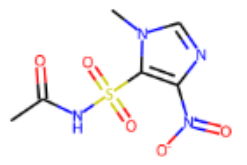 | 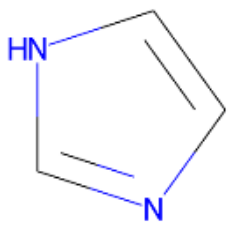 | 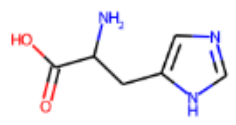 | 22.91% | 10.1021/acsami.2c13585 | |
| 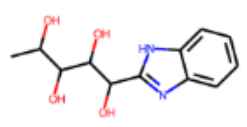 | 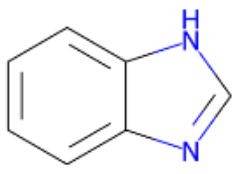 | 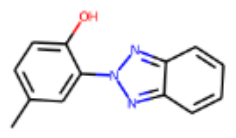 | 22.46% | 10.1016/j.jechem.2021.09.027 | |

## Experimental Validation Results

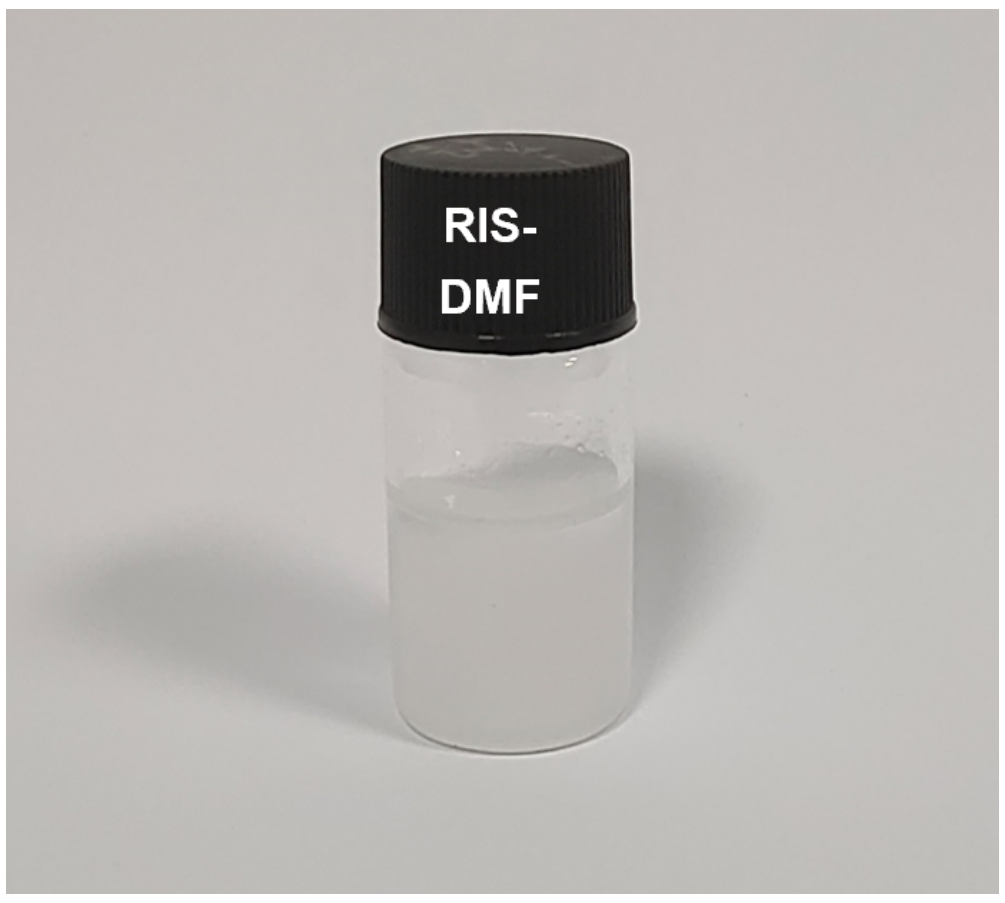

Figure S11. The picture to observe solubility of RIS in DMF.

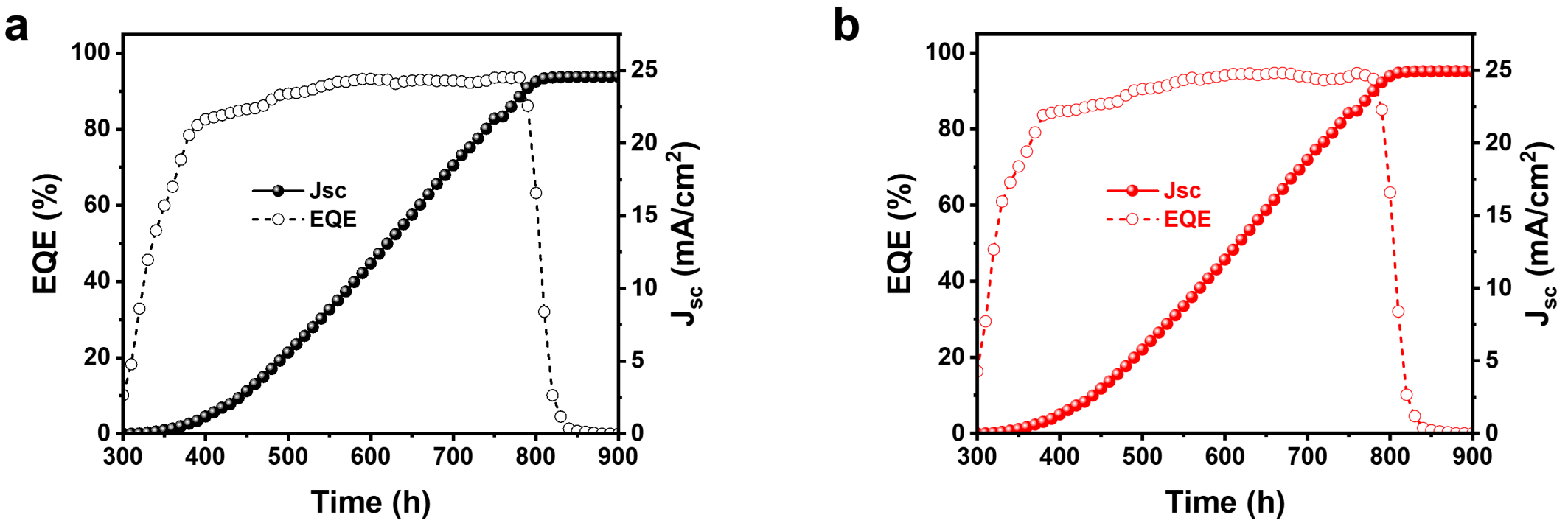

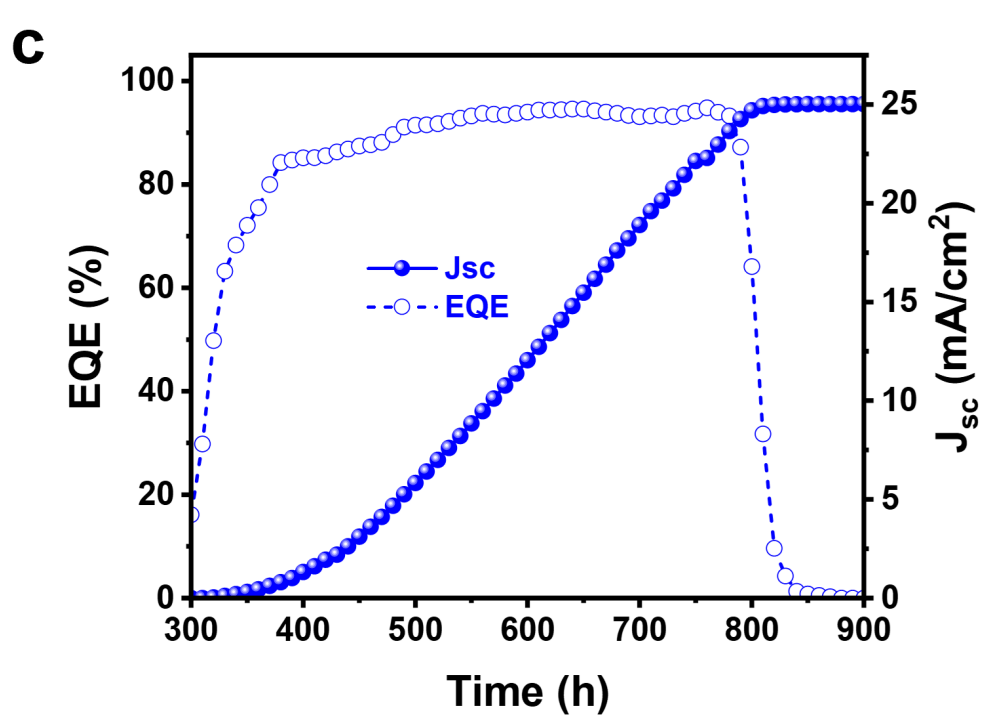

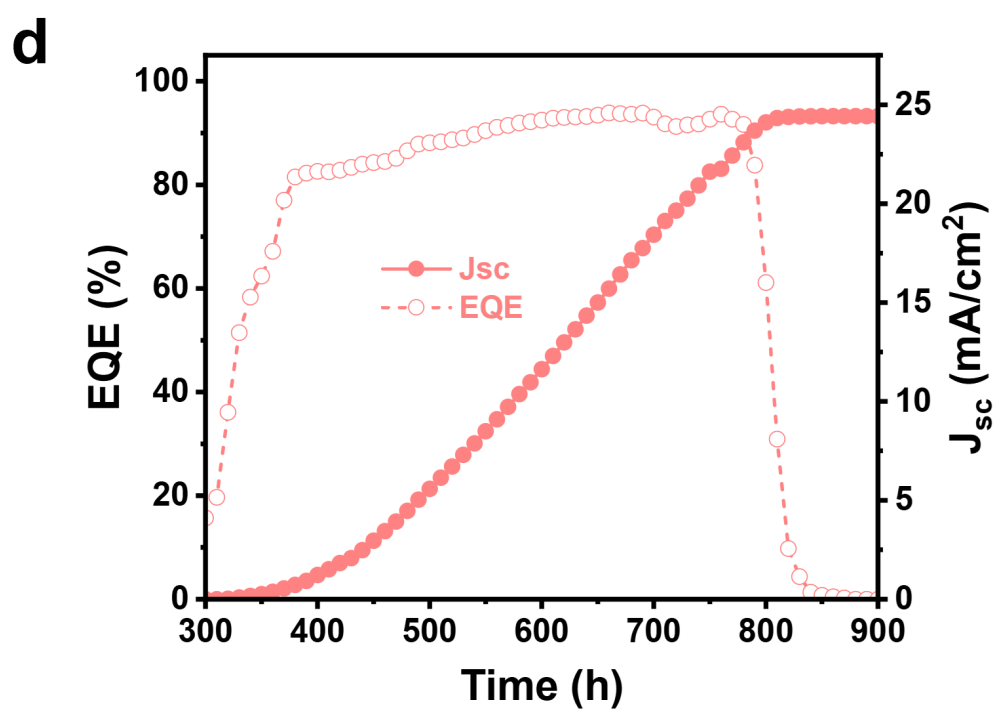

Figure S12. EQE spectra and the integrated $J_{sc}$ curves of the (a) control, (b) RIS-, (c) BTN-, and (d) NSC-treated devices.

Table S6. Photovoltaic parameters of champion PSCs with and without RIS, BTN and NSC treatment.

| **Devices** | | $V_{OC}$ **(V)** | **FF (%)** | $J_{SC}$ **(mA cm$^{-2}$)** | **PCE (%)** |
|---|---|---|---|---|---|
| RIS | Reverse | 1.164 | 83.43 | 25.11 | 24.38 |
| | Forward | 1.157 | 79.83 | 25.11 | 23.19 |
| BTN | Reverse | 1.175 | 84.58 | 25.36 | 25.20 |
| | Forward | 1.172 | 82.74 | 25.20 | 24.44 |
| NSC | Reverse | 1.123 | 81.93 | 24.68 | 22.72 |
| | Forward | 1.114 | 78.28 | 24.47 | 21.34 |
| Control | Reverse | 1.132 | 82.56 | 24.73 | 23.11 |
| | Forward | 1.126 | 77.94 | 24.54 | 21.54 |